\documentclass[journal]{IEEEtran}

\usepackage{setspace}

\usepackage{subfigure}
\usepackage{amsmath,amsfonts}
\usepackage{algorithmic}
\usepackage{algorithm}
\usepackage{array}
\usepackage{subfig}
\usepackage{textcomp}
\usepackage{stfloats}
\usepackage{url}
\usepackage{verbatim}
\usepackage{graphicx}
\usepackage{cite}
\usepackage{subfloat}
\usepackage{tikz}
\usepackage{bm}

\usepackage{color,xcolor}
\usepackage{caption}
\usepackage{pgfplots}
\usetikzlibrary{calc,arrows.meta,positioning}
\usetikzlibrary{shapes.arrows}
\usetikzlibrary{fit,backgrounds} 
\usepackage{standalone}
\usepackage[numbers,sort&compress]{natbib}
\usepackage{setspace}

\begin{document}

\title{Communication-Efficient Framework for Distributed Image Semantic Wireless Transmission}

\author{Bingyan Xie, Yongpeng Wu,~\IEEEmembership{Senior Member,~IEEE,} Yuxuan Shi, \\ Derrick Wing Kwan Ng,~\IEEEmembership{Fellow,~IEEE,} and Wenjun Zhang,~\IEEEmembership{Fellow,~IEEE}

\thanks{(Corresponding author: Yongpeng Wu.)}
\thanks{Bingyan Xie, Yongpeng Wu, and Wenjun Zhang are with the Department of Electronic Engineering, Shanghai Jiao Tong University, Shanghai 200240, China (e-mail:bingyanxie, yongpeng.wu, zhangwenjun@sjtu.edu.cn).}
\thanks{Yuxuan Shi is with the School of Cyber and Engineering, Shanghai Jiao Tong University, Shanghai 200240, China(e-mail:ge49fuy@sjtu.edu.cn).}
\thanks{Derrick Wing Kwan Ng is with the School of Electrical Engineering and Telecommunications, University of New South Wales, Sydney, NSW 2052, Australia (e-mail:w.k.ng@unsw.edu.au).}

}



\maketitle

\begin{abstract}
Multi-node communication, which refers to the interaction among multiple devices, has attracted lots of attention in many Internet-of-Things (IoT) scenarios. However, its huge amounts of data flows and inflexibility for task extension have triggered the urgent requirement of communication-efficient distributed data transmission frameworks. In this paper, inspired by the great superiorities on bandwidth reduction and task adaptation of semantic communications, we propose a federated learning-based semantic communication (FLSC) framework for multi-task distributed image transmission with IoT devices. Federated learning enables the design of independent semantic communication link of each user while further improves the semantic extraction and task performance through global aggregation. Each link in FLSC is composed of a hierarchical vision transformer (HVT)-based extractor and a task-adaptive translator for coarse-to-fine semantic extraction and meaning translation according to specific tasks. In order to extend the FLSC into more realistic conditions, we design a channel state information-based multiple-input multiple-output transmission module to combat channel fading and noise. Simulation results show that the coarse semantic information can deal with a range of image-level tasks. Moreover, especially in low signal-to-noise ratio and channel bandwidth ratio regimes, FLSC evidently outperforms the traditional scheme, e.g. about 10 peak signal-to-noise ratio gain in the 3 dB channel condition.

\end{abstract}

\begin{IEEEkeywords}
Internet-of-Things, semantic communication, distributed image transmission, hierarchical vision transformer, channel state information.
\end{IEEEkeywords}

\section{Introduction}\label{s1}
\IEEEPARstart{W}{ith} the rapid developments of the sixth-generation (6G) networks, the interaction among multiple intelligent devices has attracted intensive attention in numerous Internet-of-Things (IoT) scenarios, e.g. unmanned aerial vehicle (UAV) surveillance systems, the Internet-of-Vehicles (IoV), smart cities, etc. Correspondingly, these intelligent scenarios have posed significant demands on massive connection and transmission efficiency through the wireless networks for the huge communication and computation cost. A promising solution to these multi-node communications is the distributed data transmission, where several IoT devices only encode and transmit their own data separately while a public receiver receives and decodes all the transmitted codewords to acquire accurate messages. Through shifting the major communication cost to the receiver, cost in each IoT device becomes reasonably affordable.

\subsection{Prior Works}
The distributed data transmission aims to build a many-to-one communication system which not only satisfies accurate data transmission but also reduce the total communication cost. To address these issues, Wang et al. \cite{D-JSCC} have presented a distributed D-JSCC scheme which is able to encode and compress images from each device and jointly decodes them at the receiver for clear received images. Li et al. \cite{UAV} have used UAVs to monitor the electrical transmission lines. To save the communication cost for the target area, a Region of Interest (ROI) coding method have been proposed to effectively improve the clear reading of the target. Mital et al. \cite{stere} have focused on the stereo images and proposed a distributed source coding (DSC) method with decoder side information to further compress the raw images.

The aforementioned frameworks based on deep learning techniques though show great potential for distributed image transmission, they will underperform inevitably in some intelligent scenarios with high-resolution source input, changeable downstream tasks. This hence motivates the development of the semantic communication, which is a new paradigm paying more attention to the semantic meanings rather than the transmitted symbols. It serves as a promising paradigm to satisfy the communications among multiple intelligent devices. Conventional communications aim to transmit and recover the symbol sequences accurately at the receiver, while semantic communications focus on the transmission of meanings for specific tasks, instead of the raw data symbols. As a result, semantic communications are able to benefit from significant compression ratio and transmission efficiency \cite{sc}-\cite{zhang2022toward}.

The concept of transmission on semantic level was first proposed by Shannon and Weaver \cite{Shannon}. Over the past 70 years, this concept has been confirmed by characterizing the semantic information via logistic probability measure \cite{Carnap1952}, backgroud knowledge \cite{background}, and semantic bases \cite{zhang2022toward}. Although Jiang et al. \cite{error} have indicated that existing communications are hardly aware of the semantics based on the traditional techniques, the roll-out of the fast-growing artificial intelligence (AI) serves as the key to realize semantic communication systems. Recently, a series of semantic communication techniques have emerged with the aid of deep learning methods for handling different media, e.g. texts, images, and videos. For instance, Luo et al. \cite{luo2021autoencoder} have developed an autoencoder for text semantic transmission over Rayleigh channels. Xie et al. \cite{DeepSC} have proposed a transformer-based \cite{Vaswani} deep learning enabled semantic communication (DeepSC) framework. Also, Huang et al. \cite{huang2021} have exploited an improved generative adversarial network (GAN) for semantic coding and image transmission. Zhou et al. \cite{fast} have taken the time-varying channel into consideration and thus proposed a feature arrangement scheme according to future channel state information (CSI) for semantic communication. Hu et al. \cite{robust} have designed the masked vector quantized-variational autoencoder (VQ-VAE) for combating the semantic noise in image transmission. Besides, Jiang et al. \cite{videos} have established a wireless semantic network, called semantic video conferencing (SVC), for enabling the effective transmission of crucial elements contained in the video. 

In light of the advantages of semantic communications on saving bandwidth resources, there are some initial works applying this emerging paradigm to distributed data transmission scenarios. For example, Xie et al. \cite{lite} have focused on a simple distributed text transmission scenario and proposed a lite distributed semantic communication (L-DeepSC) system for IoT edge devices. Besides, Tong et al. \cite{globe2021} have combined federated learning (FL) \cite{McMahan} with semantic communication, where FL enables the semantic integration of multiple users. Furthermore, Shi et al. \cite{shi2021semantic} have introduced the models, architectures, and challenges for nowadays semantic communication and then proposed a federated edge intelligence (FEI)-based architecture for the realization of semantic-aware networking.

\subsection{Motivation and Contributions}

Although the above works \cite{lite}-\cite{globe2021}, \cite{shi2021semantic} utilize semantic communication into the distributed transmission scenarios to save communication resources, the much huger data streams in image transmission call for the implementation of more communication-efficient frameworks. Then, they seldom consider task adaptation, which means the inflexibility in handling changes in the downstream tasks. Also, some existing works view the data transmission of each user as totally isolated procedure, e.g. \cite{lite}, thus not fully improving the semantic extraction and task performance through the correlation among users. Moreover, the above works mainly consider the additive Gaussian noise or stable channel fading for each user, which is not applicable to practical varying wireless channels. Although existing works \cite{luo2021autoencoder}, \cite{fast} study semantic transmission in time-varying channels with time property, their results are designed for single-antenna devices and are not applicable to emerging multiple-input multiple-output (MIMO) wireless transmission with both time and spatial properties.

In this paper, we focus on a multi-node surveillance scenario where several IoT devices like cameras or sensors monitor a single scene while the images are encoded and transmitted through noisy channels and then processed for intelligent downstream tasks. Thus, we propose a federated learning-based semantic communication (FLSC) framework. It conducts efficient distributed image transmission based on the semantics for specific downstream tasks. The huge communication cost can be saved through transmitting required semantics while task adaptation is satisfied by the layered structure. More precisely, a hierarchical feature extractor is designed for images, which performs semantic enhancement and is adaptive to different downstream tasks. We further propose a semantic-aware FL algorithm for global optimization, where semantic information from each device are aggregated to improve task performances, e.g. better total classification accuracy for image classification or a complete panorama to present the full monitored scene in image reconstruction task. To further extend the FLSC into realistic varying channel conditions, a CSI-based MIMO transmission module including the singular value decomposition (SVD)-based MIMO precoder and detector, and the two-step channel estimator is considered into the FLSC for combating the channel variation. The main contributions of this work are summarized as follows:

\begin{enumerate}
\item{We propose a novel federated learning-based semantic communication (FLSC) framework, which combines FL with semantic communication to ensure effective distributed image transmission with IoT devices. Different from the traditional FL algorithms, we propose a semantic-aware FL algorithm which aggregates both the global weights and the task results for global aggregation. The isolated FL link of each user is able to exchange semantic knowledge and share the task results to improve final task performances.}
\item{We design a hierarchical vision transformer (HVT)-based encoder and a task-adaptive translator for semantic extraction and understanding. The HVT-based encoder extracts the task-specific semantics for transmission, thus saving bandwidth resources to a great extent. For the flexibility of implementing different tasks, the HVT structure extracts semantic information from coarse to fine with different depths of layers. Additionally, an improved module named spatial reduction local self-attention (SRLSA) module is embedded in the HVT for complexity reduction and semantic enhancement. At the receiver, the translator exploits task-specified modules to interpret the semantic information for obtaining the final results.}	
\item{We consider the MIMO fading channels into the FLSC to further extend the framework into more realisitc varying channel conditions. To overcome fading in the MIMO channels, a two-step channel estimation strategy is proposed. Least square (LS) channel estimation is introduced at the receiver to acquire the coarse CSI first. After that, a U-channel estimator is proposed and embedded at the receiver to purify the coarse CSI from the LS method. Then the purified fine CSI is fed back to the transmitter. With the timely estimated CSI, the SVD-based MIMO precoding and detection is able to significantly alleviate channel fading.}
\end{enumerate}

This paper is organized as follows. In Section II, we introduce the system model and describe the overall structure of the FLSC framework into three levels, namely semantic, transmission, and application level. In Section III, we show the details of the modules according to the three levels and present the task-related designs for image classification and image reconstruction tasks, respectively. In Section IV, we provide numerical results to demonstrate the superiority of our FLSC compared with other candidates, especially in the aspects of limited bandwidth resources, and low signal-to-noise ratio (SNR) for various tasks. Finally, Section V concludes the paper and presents some future directions.

Notational Conventions: $\mathbb{R}$ and $\mathbb{C}$ refer to the real and complex number sets, respectively. $\mathcal{CN}\left (\mu, \sigma^2 \right)$ denotes the complex form of a Gaussian function with mean $\mu$ and variance $\sigma^2$. Finally, $\left(\cdot\right)^{H}$ denotes the Hermitian.

\section{System Model}
In this section, we first introduce the scenario of distributed image wireless transmission and the proposed FLSC framework. Following Weaver and Shannon \cite{Shannon}, the framework can be interpreted into three levels: semantic, transmission, and application level. The semantic-aware FL method is also provided in this section.

\begin{figure*}[hbtp]
	\centering
	\includegraphics[width=6.6in]{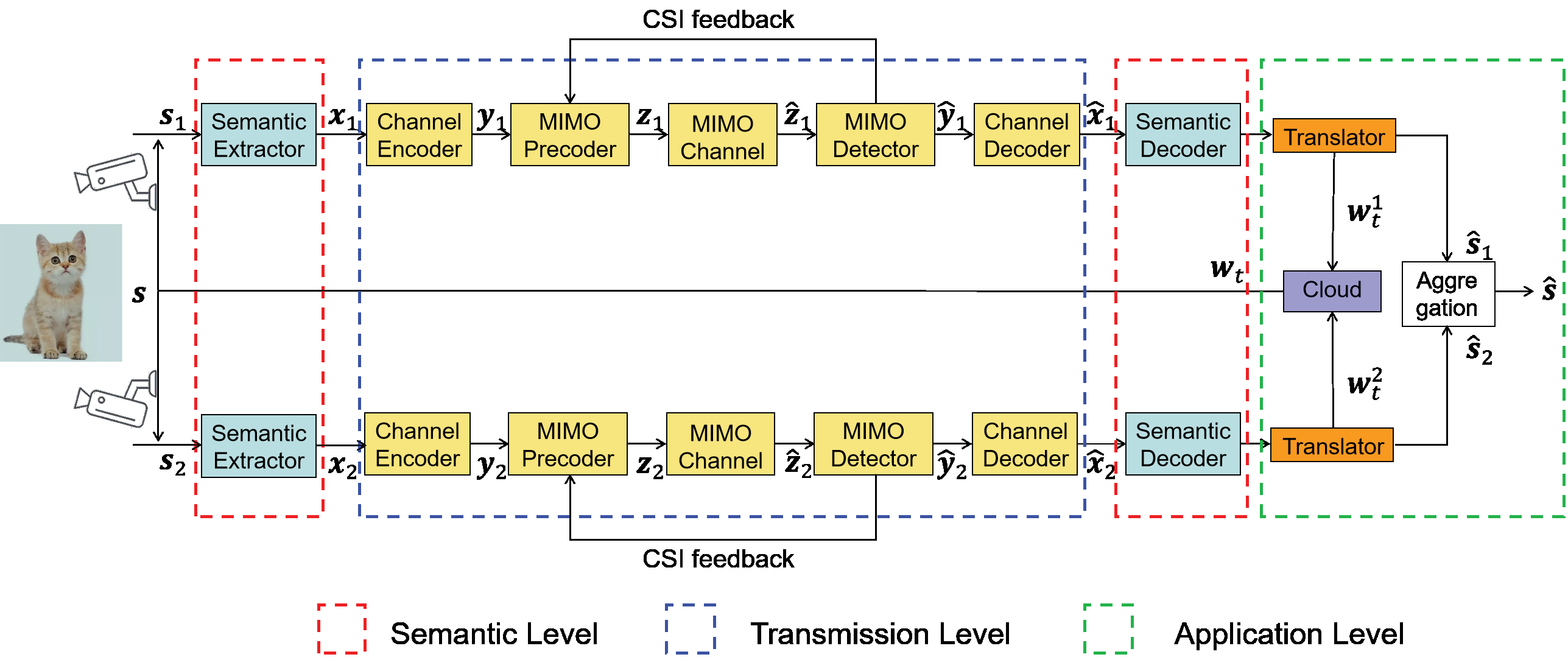} 
	\caption{The proposed federated learning-based semantic communication framework for distributed image transmission with IoT devices in MIMO fading channels ($N=2$).}
	\label{fig_1}
\end{figure*}

We consider the following scenario in Fig. \ref{fig_1}. There are $N$ cameras or sensors at different locations taking pictures of the same object or scene, $\bm{s}$. Then, these different images are encoded and transmitted through noisy channels to a common receiver for multiple downstream tasks. To describe the correlation between the images from the adjacent cameras, we introduce a parameter called image overlap ratio $\delta\in[0,1]$. The captured image $\bm{s}_i$ at the $i$-th device is then sent into the semantic encoder, where $\bm{s}_i \in\mathbb{R}^{H\times W\times C}$, for $i=1,2,...,N$. Here $H\times W\times C$ denotes the size of the original image. At the transmitter, a semantic extractor, $S_{\bm{\alpha}}(\cdot): \mathbb{R}^{H\times W\times C} \mapsto \mathbb{R}^{C_{\mathrm{H}}}$, extracts image semantics $\bm{x}_i\in\mathbb{R}^{C_{\mathrm{H}}}$ and a channel encoder, $F_{\bm{\beta}}(\cdot): \mathbb{R}^{C_{\mathrm{H}}} \mapsto \mathbb{R}^{C_{\mathrm{L}}}$, outputs the codewords $\bm{y}_i\in\mathbb{R}^{C_{\mathrm{L}}}$ with parameters $\bm{\alpha}$ and $\bm{\beta}$, respectively, where $C_{\mathrm{H}}$ refers to the sequence length of the extracted semantic information while $C_{\mathrm{L}}$ refers to the length of the codewords. Finally, the MIMO precoder, $P_{\mathrm{r}}(\cdot): \mathbb{R}^{C_{\mathrm{L}}} \mapsto \mathbb{R}^{N_{\mathrm{T}}\times \frac{C_{\mathrm{L}}}{N_{\mathrm{T}}}}$, preprocesses $\bm{y}_i$ into $\bm{z}_i \in \mathbb{R}^{N_{\mathrm{T}}\times \frac{C_{\mathrm{L}}}{N_{\mathrm{T}}}}$ in order to combat the channel fading. The above process can be expressed as
\begin{align}
	\bm{s}_i\xrightarrow[]{S_{\bm{\alpha}}(\cdot)}\bm{x}_i\xrightarrow[]{F_{\bm{\beta}}(\cdot)}\bm{y}_i\xrightarrow[]{P_{\mathrm{r}}\left(\cdot\right)}\bm{z}_i.
\end{align}

After the MIMO precoder, the codewords $\bm{z}_i$ for the $i$-th device are sent to the channel. We consider the MIMO fading channel, in which each user owns the different CSI independently. The MIMO channel matrix is denoted as $\bm{h}\in\mathbb{C}^{N_{\mathrm{R}}\times N_{\mathrm{T}}}$, whose component follows the distribution $\mathcal{CN}(0,1)$. The received codewords from the $i$-th device can be expressed as
\begin{align}
	\label{11}
	\hat{\bm{z}_i}=\bm{h}_i\bm{z}_i+\bm{n}_i,     
\end{align}
where $\bm{n}_{i} $ is the Gaussian channel noise vector whose component has zero mean and covariance $\sigma_{n_i}^{2}$.

The decoder is composed of a MIMO detector, $D_{\mathrm{e}}(\cdot): \mathbb{R}^{N_{\mathrm{R}}\times \frac{C_{\mathrm{L}}}{N_{\mathrm{T}}}} \mapsto \mathbb{R}^{C_{\mathrm{L}}}$, a channel decoder, $D_{\bm{\theta}}(\cdot): \mathbb{R}^{C_{\mathrm{L}}} \mapsto \mathbb{R}^{C_{\mathrm{H}}}$, a semantic decoder, and a task-oriented translator. Working together with the precoder $P_{\mathrm{r}}(\cdot)$, the MIMO detector $D_{\mathrm{e}}(\cdot)$ utilizes the estimated CSI to recongnize the received sequence $\bm{\bm{\hat{z}}}_i$ as $\bm{\bm{\hat{y}}}_i$. The channel decoder obtains $\bm{\bm{\hat{x}}}_i$ from the noisy codewords $\bm{\bm{\hat{y}}}_i$, where $\bm{\bm{\hat{y}}}_i$ and $\bm{\bm{\hat{x}}}_i$ have the same dimensions as $\bm{y}_i$ and $\bm{x}_i$, respectively. Then, the semantic decoder coarsely translates the semantics rectified by the channel decoder. For task adaptation and complexity reduction, the task-oriented translator interprets the semantic information into different results $\hat{\bm{s}}_i$ according to the task difficulties. For simplicity, we combine the semantic decoder with the task-oriented translator and denote the overall task-adaptive decoder, $G_{\bm{\eta}}(\cdot): \mathcal{\hat{X}}_i \mapsto \mathcal{\hat{S}}_i$, where $\mathcal{\hat{X}}_i$ and $\mathcal{\hat{S}}_i$ are the sets of $\bm{\hat{x}}_i$ and $\bm{\hat{s}}_i$, respectively. The above process can be expressed as
\begin{align}
	\hat{\bm{z}}_i\xrightarrow[]{D_{\mathrm{e}}(\cdot)}\hat{\bm{y}}_i\xrightarrow[]{D_{\bm{\theta}}(\cdot)}\hat{\bm{x}}_i\xrightarrow[]{G_{\bm{\eta}}(\cdot)}\hat{\bm{s}}_i.
\end{align}

Following the above process in equations (1)-(3), each independent point-to-point semantic communication link receives the task result $\bm{\hat{s}}_i$, which can be denoted as
\begin{align}
	\hat{\bm{s}}_i = G_{\bm{\eta}}(D_{\bm{\theta}}(D_{\mathrm{e}}(P_{\mathrm{r}}(F_{\bm{\beta}}(S_{\bm{\alpha}}(\bm{s}_i))), \bm{h}_i, \bm{n}_i))),
\end{align}
where for each user $i$ in FLSC, the above defined $G_{\bm{\eta}}, D_{\bm{\theta}}, F_{\bm{\beta}}$, and $S_{\bm{\alpha}}$ have different model weights for the later FL aggregation.

To improve the transmission performance of the whole system, at the current global training epoch $t$, the model weights of each device, $\bm{w}_{t}^i$, are sent to the cloud server for constructing a global model with weights, $\bm{w}_{t}=(\bm{\alpha},\bm{\beta},\bm{\theta},\bm{\eta})$, which are the weight average from all the local devices. Different from the traditional FL method, at the same time, all the results, $\bm{\hat{s}}_1,\cdots,\bm{\hat{s}}_N$, from each device aggregate together as $\bm{\hat{s}}$ to jointly complete the specific tasks. At the next global training epoch $t+1$, the local model weights, $\bm{w}_{t+1}^1, \bm{w}_{t+1}^1, \cdots, \bm{w}_{t+1}^N$, are replaced by the same global model weights. The weights update process can be described as
\begin{align}
	\label{2}
	\bm{w}_{t+1} = \frac{1}{N}{\textstyle \sum_{k=1}^{N}} \bm{w}_{t}^{k}.
\end{align}

Considering different tasks, we define the task-specified function as $\varphi(\cdot):\mathcal{S}_i\mapsto \mathcal{\hat{S}}_i$, where $\mathcal{S}_i$ is the set of $\bm{s}_i$. More specifically, $\varphi(\bm{s})$ is the true label in image classification or a benchmark image in the image reconstruction task. We denote the FLSC aggregation process $\Psi_{\bm{w}}(\cdot): \mathcal{\hat{S}}_1\times \cdots \times \mathcal{\hat{S}}_N \mapsto \mathcal{\hat{S}}$, with $w$ represents the final global training epoch weights. Through a series of processes, including semantic-level coding, transmission-level MIMO communication, and application-level task adaptation and aggregation, each user utilizes their own source images and communication link to acquire task results. Then, these independent task results can be aggregated to obtain the final global results. The overall distributed image transmission process can be formulated as

\begin{align}
	\label{19}
	\hat{\bm{s}}=\Psi_{\bm{w}}(\bm{\hat{s}}_1,\cdots,\bm{\hat{s}}_N).
\end{align}

Based on this semantic-aware FL algorithm, each user is able to exchange semantic information with each other, and hence obtains admirable task performances. Fig. \ref{fig_1} shows the federated learning-based semantic communication framework with $N=2$ devices.

\section{Three-level Semmantic Communication System}
In this section, we present the design details for each level. Additionally, the application level of the task-related designs, including the semantic translator, the result aggregation method, and the semantic loss function are given. Finally, we provide the corresponding training algorithms for the FLSC with different tasks.

\subsection{Semantic Level}

At the semantic level of the FLSC framework, we present the designs of the semantic extractor, $S_{\bm{\alpha}}(\cdot)$, and the semantic decoder. The latter is a component of the task-adaptive translator, $G_{\bm{\eta}}(\cdot)$. In this level, to extract different scales of features for specific tasks, we propose a hierarchical vision transformer (HVT) to extract the semantic information $\bm{x}_i$ from the raw image $\bm{s}_i$ at the $i$-th device. While for reducing the computation cost, the asymmetric decoder structure called deep transpose convolution-based neural network (DTCNN) is introduced to reconstruct the original image through received semantic information $\bm{\bm{\hat{x}}_i}$.                   
\subsubsection{Semantic Extractor}

Consider the image-relevant tasks, the transformer-based models are able to extract global semantic information better than CNNs \cite{CNN}, due to the self-attention mechanism which ensures long-term memory for each image patch. \cite{Dosovitskiy} Hence, we propose an HVT-based encoder for semantic extraction in the FLSC framework. Moreover, a new mechanism called spatial reduction local self-attention (SRLSA) is adopted for computational complexity reduction and semantic enhancement. The details are presented as follows.

\begin{figure}[htbp]
	\centering
	\includegraphics[width=3.4in]{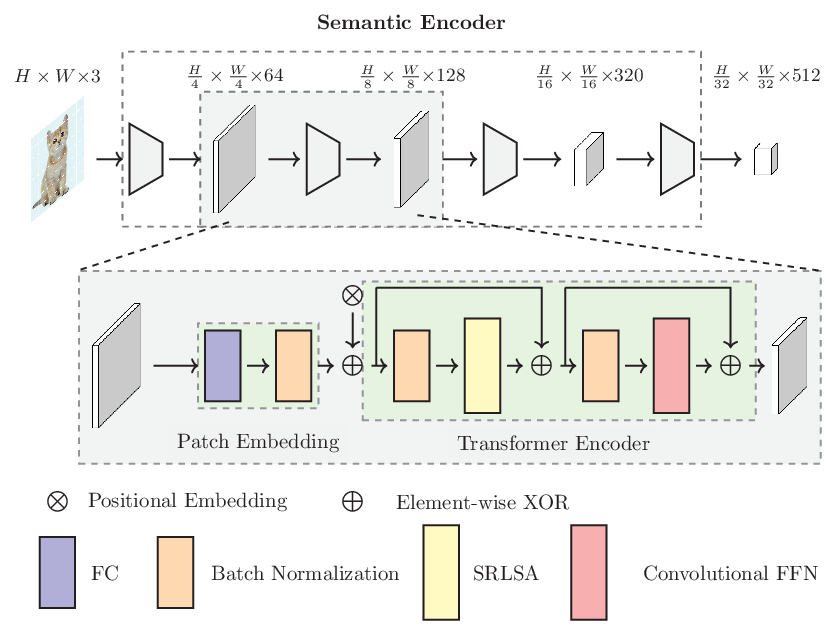}
	\caption{The structure of a hierarchical vision transformer (HVT).}
	\label{fig_2}
\end{figure}

The semantic extractor, HVT, is originated from the classical pyramid structure that piles several ViT structures up together as a hierarchical extractor \cite{Wang1}-\cite{liu2021semantics}. As shown in Fig. \ref{fig_2}, HVT is a four-stage network, where each stage is composed of the patch embedding, positional encoding, and transformer encoder modules. The output channel numbers of different stages are 64, 128, 320, 512, respectively.
In each stage, semantic information can be extracted and further refined. Hence, coarse-to-fine semantic information can be obtained through different stages. In the proposed FLSC framework, the semantic features from the superficial stages are enough for some image-level tasks like image classification as semantics from the whole image are not needed, while the features from the deeper stages are necessary for some pixel-level tasks, e.g. image reconstruction, since full sematic information is required to accurately reconstruct images at the receiver. Extracting and reshaping the semantic information into different scales can provide the flexibility for selecting the scalable semantics in different tasks.

\begin{figure}[htbp]
	\centering
	\includegraphics[width=0.34\textwidth]{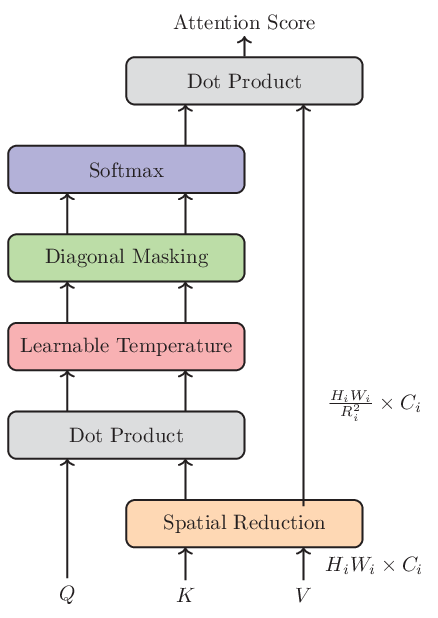}
	\caption{The structure of spatial reduction local self-attention (SRLSA).}
	\label{fig_3}
\end{figure}

A series of improvements are added in the HVT. First, we utilize the overlapped patch embedding method \cite{Wang2} by adding extra zero-padding embeddings to expand the feature map. The overlap among adjacent patch embeddings helps the semantic extractor better refine the semantic relevance among different patches. Next, we also present an improved module, called spatial reduction local self-attention (SRLSA), which is adopted for computation complexity reduction and semantic enhancement in the HVT encoder instead of the conventional multi-head attention (MHA) module. The SRLSA combines spatial reduction attention (SRA) \cite{Wang1} mechanism with local self-attention (LSA) \cite{Lee} mechanism as shown in Fig. \ref{fig_3}, which can both help extract semantic information better and reduce the computation cost. 

The LSA module consists of two core parts, namely diagonal mask and learnable temperature scaling, respectively. The attention scores, $\bm{A}_t \in \mathbb{R}^{P\times P}$, use two mapping matrices, called query $\bm{Q}\in \mathbb{R}^{P \times C_{\mathrm{H}}}$ and key $\bm{K}\in \mathbb{R}^{P \times C_{\mathrm{H}}}$ to form the square matrices to show the semantic correlations among overall $P$ numbers of different patch embeddings. Then, the third mapping matrix, $\bm{V}\in \mathbb{R}^{P \times C_{\mathrm{H}}}$, measures values of all the patch embeddings and is multiplied by the attention scores, $\bm{A}_t$. The attention weight, $\bm{A}_w\in \mathbb{R}^{P\times C_{\mathrm{H}}}$, is able to capture the important extent for different patch embeddings. We assume $\bm{A}_w$ as the required semantic information and $\bm{A}_t$ is the reflection of the whole tendency of semantics $\bm{A}_w$. Note that higher $\bm{A}_t$ represents the closer correlation between the two arbitrary patch embeddings. The diagonal mask sets the attention scores in the diagonal positions as -$\infty$ to distract the attention from itself to other embeddings, thus enhancing the semantic interaction among different segmented patches. The diagonal mask attention scores, $\bm{A}_{t}^{M} \in \mathbb{R}^{P\times P}$, are stated as 
\begin{align}
	\label{4}
\bm{A}_{t(i,j)}^{M}\left ( \bm{m} \right ) =\left\{\begin{matrix} 
	\bm{A}_{t(i,j)}\left ( \bm{m} \right ), \qquad  \left (i\ne j  \right ), \\  
	-\infty,  \qquad \quad \left (i= j  \right ),
\end{matrix}\right. 
\end{align}
where $\bm{A}_{t(i,j)}$ is the attention score between the $i$-th and $j$-th patch embeddings, $\bm{m}$ is the mapping matrix $\bm{Q}$ or $\bm{K}$.

The learnable temperature scaling adds a new learnable parameter, called temperature $\tau_{\mathrm{L}}$, for softmax function $\varPhi(\cdot)$ to adaptively scale the attention scores to a sharper distribution during the training process. For the SRA, we perform the spatial reduction for $\bm{K}$ and $\bm{V}$ while the remained mapping, $\bm{Q}$, unchanged. The computation of the spatial reduction is defined as
\begin{align}
	\label{8}
S_{\mathrm{R}}\left ( \bm{x} \right )=N_{\mathrm{o}}\left (R_{\mathrm{e}}\left (\bm{x},R_{i}\right) \cdot W^{S}\right),   
\end{align}
where the initial size of input $\bm{x}$ for the $i$-th layer in HVT is $H_{i}\times W_{i}\times C_{i}$ and the convert function is defined as $R_{\mathrm{e}}(\cdot): \mathbb{R}^{H_{i}\times W_{i}\times C_{i}} \mapsto \mathbb{R}^{\frac{H_{i}W_{i}}{R_{i}^{2}}\times\left (R_{i}^{2}C_{i}\right)}$. Then by multiplying extra spatial reduction matrix $W^{S}\in \mathbb{R}^{\left (R_{i}^{2}C_{i}\right)\times C_{i}}$, the dimension of $\bm{K}$ and $\bm{V}$ can be reduced to $R_{i}^{2}{\tiny {\tiny }}$ times compared with the initial self-attention ones. Finally, $N_{\mathrm{o}}(\cdot)$ performs the normalization function.

The final SRLSA attention weights, $\bm{A}_w$, can be formulated as

\begin{align}
	\label{9}
\bm{A}_w\left(\bm{Q},\bm{K},\bm{V}\right)=\varPhi\left(\frac{\bm{A}_t^{M}\left(\bm{Q}\cdot S_{\mathrm{R}}\left (\bm{K}\right)^{T}\right )}{\tau_{\mathrm{L}}}\right) \cdot S_{\mathrm{R}}\left (\bm{V}\right).  
\end{align}

Herein we intepret the attention weights, $\bm{A}_w$, as the semantic information extracted by the HVT. Note that for various tasks, the necessary semantics are different. Take the image classification as an example, an extra class (CLS) embedding is set to evaluate the correlation among itself and other embeddings. Thus, the transmitted semantics are only the attention weights of CLS embedding. However, image reconstruction aims to recover the whole original images. Therefore, the full attention weights are assumed as the semantic information prepared for transmission. The distribution of $\bm{A}_t$ is also an evaluation index for the performance of semantic information extraction since it can reflect how important different patch embeddings are. By combining the above modules, the SRLSA makes the distribution of attention scores sharper than the original MHA, which means it retains the more robust features for the later computed semantics $\bm{A}_w$.

\subsubsection{Semantic Decoder}
The semantic decoder is a selective part in the joint task-adaptive translator, $G_{\bm{\eta}}(\cdot)$, and is used for semantics recovery and tasks execution. In light of the complexity reduction, we adopt the asymmetric structure where the encoder is a transformer and the decoder is a deep transpose convolution-based neural network (DTCNN). As illustrated in \cite{MAE}, the asymmetric structure does not reduce the performance on final task results but turns more lighted-weighted compared with the symmetric codec based on transformers. The DTCNN is mainly modified by transpose convolution and the pixel shuffle module \cite{Shi}.

The first three layers consist of a transpose convolution layer, an activation layer, and a batch normalization layer, respectively. Take the 32$\times$32$\times$3 CIFAR10 dataset as an example: After being purified by the semantic extractor, the scale of the initial encoded semantic information is 2$\times$2. The transpose convolution layer uses kernel size 4, stride 2, and padding size 1 structure to expand the scale of the semantic information map as 4$\times$4, 8$\times$8, and 16$\times$16, respectively, while the final channel number is reduced to 12. We choose the Gaussian error linear units (GELU) \cite{GELU} as the activation function, which enjoy better performance than the common rectified linear unit (ReLU) function. The final layer is composed of a pixel shuffle module and a sigmoid activation function. The stride of pixel shuffle is set as 0.5 so that the channel number is reduced to 3 and the semantic information map is expanded as 32$\times$32. The sigmoid function maps the decoded semantic information to the range of [0,1]. 

\subsection{Transmission Level}

\begin{figure*}[htbp]
	\centering
	\includegraphics[width=4.5in]{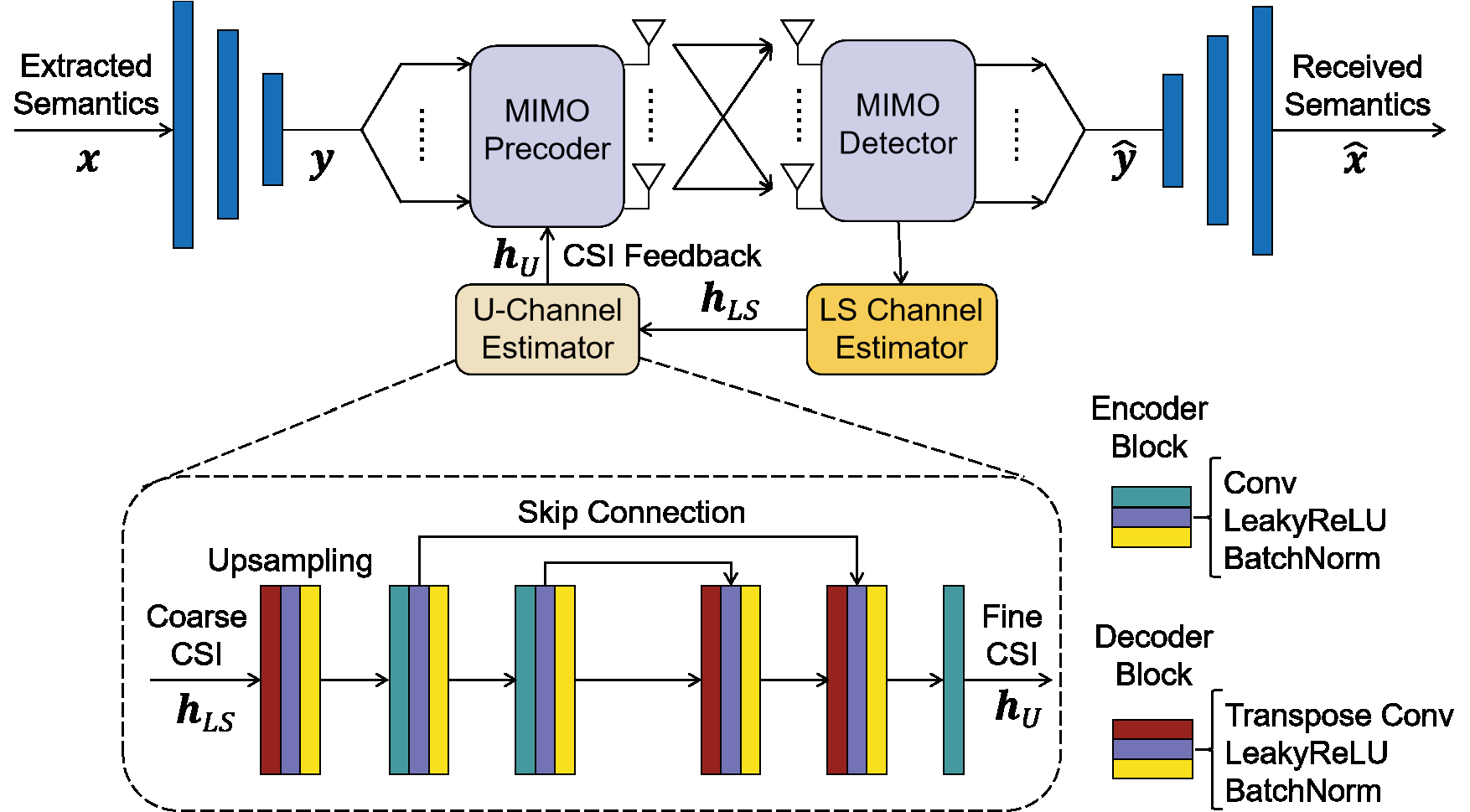}
	\caption{CSI-based MIMO semantic transmission model.}
	\label{fig_20}
\end{figure*}

At the transmission level, to further improve the transmission efficiency, the MIMO techniques are introduced for semantic transmission in Fig. \ref{fig_20}. The CSI-based MIMO semantic transmission model consists of the MIMO channel, the MIMO precoder and detector, and a two-step CSI estimator.

After compressing and further extracting the semantic information, then $\bm{y}$ is divided into $N_{\mathrm{T}}$ streams to fit the MIMO channel. For the MIMO precoding, we adapt the SVD methods, which can be formulated as
\begin{align}
	\label{15}
	\bm{h}_\mathrm{U}=\bm{U}\bm{\Lambda} \bm{V}_{\mathrm{p}}^{H},     
\end{align}
where $\bm{h}_\mathrm{U} \in \mathbb{C}^{N_{\mathrm{R}} \times N_{\mathrm{T}}}$ refers to the latter estimated CSI by U-Channel Estimator, $\bm{U} \in \mathbb{C}^{N_{\mathrm{R}} \times N_{\mathrm{R}}}$ and $\bm{V}_{\mathrm{p}} \in \mathbb{C}^{N_{\mathrm{T}} \times N_{\mathrm{T}}}$ are both decomposed unitary matrices, $\bm{\Lambda} \in \mathbb{C}^{N_{\mathrm{R}} \times N_{\mathrm{T}}}$ is the diagonal matrix.

To preprocess the transmitted data, the precoded semantics are shown as
\begin{align}
	\label{16}
	\bm{z}=P_{\mathrm{r}}\left(\bm{y}\right)=\bm{V}_{\mathrm{p}}\bm{y}.     
\end{align}

After going through the MIMO fading channel illustrated in (\ref{11}), we conduct the MIMO detection in accordance with the SVD precoding
\begin{align}
	\label{17}
	\bm{\bm{\hat{y}}}=D_{\mathrm{e}}\left(\bm{z}\right)=\bm{\Lambda} ^{-1}\bm{U}^H\bm{\bm{\hat{z}}}=\bm{\Lambda} ^{-1}\bm{U}^H\bm{h}\bm{V}_{\mathrm{p}}\bm{y}+\bm{\Lambda} ^{-1}\bm{U}^\textsc{H}\bm{n}.    
\end{align}

Since the practical channel matrix $\bm{h}$ can also conduct the SVD, if the predicted CSI $\bm{h}_\mathrm{U}$ is accurate enough, the received $\bm{\hat{y}}$ is able to significantly alleviate the effect of MIMO fading and retain the original extracted semantic information.

To acquire and exploit the accurate CSI, We estimate the CSI at the receiver and send the CSI feedback to the transmitter. For the channel estimation, we propose a two-step method, in which the coarse CSI, $\bm{h}_\mathrm{LS}$, is estimated by the least square (LS) \cite{LS} channel estimation method and the fine CSI, $\bm{h}_\mathrm{U}$, is predicted by the U-channel estimator \cite{U-Net}. The $\bm{h}_\mathrm{LS}$ is provided by the transmitted and received pilots.
\begin{eqnarray}
	\label{18}
	\hat{\bm{\Gamma}}&=&\bm{h}\bm{\Gamma}+\bm{n}, \\
	\bm{h}_\mathrm{LS}&=&\hat{\bm{\Gamma}}\bm{\Gamma}^{-1}, \\
	\bm{h}_\mathrm{U}&=&U_{\bm{\Upsilon}}\left (\bm{h}_\mathrm{LS}\right), 	     
\end{eqnarray}

where $\bm{\Gamma} \in \mathbb{R}^{N_{\mathrm{T}}\times N_{\mathrm{T}}}$ and $\bm{\hat{\Gamma}} \in \mathbb{R}^{N_{\mathrm{R}}\times N_{\mathrm{T}}}$ refer to the transmitted and received pilots, respectively. $U_{\bm{\Upsilon}}\left(\cdot\right)$ stands for the U-channel estimator.

The U-channel estimator, $U_{\bm{\Upsilon}}\left(\cdot\right)$, which adapts the U-Net structure \cite{U-Net}, is used for refining the coarse estimated CSI through the LS method to the fine CSI for the MIMO precoding and detection. It has two main blocks, namely encoder block and decoder block. The former consists of a convolution layer, a leakyReLU activation layer, and a Batch Normalization layer while the latter is made up of a transposed convolution layer, a leakyReLU activation layer, and a Batch Normalization layer. The $U_{\bm{\Upsilon}}\left(\cdot\right)$ first upsamples $\bm{h}_\mathrm{LS}$ to a larger feature map for the prepocessing. Then the encoder-decoder structure is applied through a series of encoder and decoder blocks. It is worth mentioning that several skip connections are added where the feature maps of encoder blocks and decoder blocks are combined by concatenating layers to retain pixel level details at different resolutions. As a result, $U_{\bm{\Upsilon}}\left(\cdot\right)$ enables the finer channel estimation than the straightforward encoder-decoder structure.

Although the two-step channel estimation can acquire fine CSI, there still exists some distortion compared to the real CSI, $\bm{h}$. When under low SNR conditions, such distortion expands a lot which brings difficulty for the $G_{\bm{\eta}}(\cdot)$ to translate semantic for the latter tasks. For the further protection of the semantic codewords under MIMO fading channels, the autoencoder-based channel encoder, $F_{\bm{\beta}}(\cdot)$, and decoder, $D_{\bm{\theta}}(\cdot)$, are also introduced in the transmission level. The channel encoder is a multi-layer structure, where the dimension of the first layer is $C_{\mathrm{H}}$ while the last layer is linear with dimension, $C_{\mathrm{L}}$, adjustable according to the channel bandwidth ratios. The channel decoder is the inverse of the encoder except the final activation layer, which is substituted by sigmoid instead of ReLU. After extracting the needed semantics, the channel encoder then maps the semantics $\bm{x}$ to transmitted sequence $\bm{y}$ which better fits the wireless transmission. Before translating the semantics, the channel decoder further purifies the received codewords. In this way, efficient semantic transmission over MIMO fading channel is enabled.

For offering a higher stability to the varying wireless MIMO channels, the U-channel estimator is taken as an independent part to be pretrained and then embedded into our FLSC. The pretained method is shown in Algorithm \ref{alg1}. A large sample set $\bm{H}$ containing large amount of MIMO CSI is proposed as raw data for the $U_{\bm{\Upsilon}}\left(\cdot \right)$. For each training step, a batch $\bm{a}\in \bm{H}$ is first selected. Then, to simulate the coarse CSI $\bm{h}_\mathrm{LS}$ estimated by the LS method, the stable pilot sequence, $\bm{\Gamma}$, transmits through the MIMO fading channel according to the sampled batch. With the received pilot sequence $\bm{\hat{\Gamma}}$, the LS method helps to obtain the coarse CSI $\bm{h}_\mathrm{LS}$. After that, the $\bm{h}_\mathrm{LS}$ is fed into the $U_{\bm{\Upsilon}}\left(\cdot \right)$ to get purified CSI $\bm{h}_\mathrm{U}$. The mean square error (MSE) loss function for training the $U_{\bm{\Upsilon}}\left(\cdot \right)$ is

\begin{align}
	\label{30}
	L_{\mathrm{MSE}}=\frac{1}{N}\sum_{i=0}^{n} ||\bm{h}_\mathrm{U}-\bm{\hat{a}}||^{2},
\end{align}
where $N$ is the number of elements in MIMO fading channel matrix.

\begin{algorithm}[htbp]
	\caption{Pretrain Method for the U-Channel Estimator}\label{alg:alg3}
	\begin{algorithmic}
		\STATE 
		$\textbf{Input:}$ Channel SNR, Pliot sequence $\bm{\Gamma}$, and Sample sets of 
		\STATE \hspace{1.1cm}MIMO channel fading matrix $\bm{H}$		
		
		$\textbf{Output:}$ The U-channel estimator $U_{\bm{\Upsilon}}\left(\cdot \right)$	
		
		\STATE \hspace{0.5cm}$ \textbf{} $
		
		\STATE \hspace{0.5cm}1. Take a batch $\bm{a}$ from the sample sets $\bm{H}$
		\STATE \hspace{0.5cm}2. Let the stable pilot sequence $\bm{\Gamma}$ faces with generated 
		\STATE \hspace{0.9cm}MIMO fading channels:
		\STATE \hspace{0.9cm}$\hat{\bm{\Gamma}}=\bm{a}\bm{\Gamma}+\bm{n}$
		\STATE \hspace{0.5cm}3. Adapt LS to obtain the coarse CSI: $\hat{\bm{a}}=\hat{\bm{\Gamma}}\bm{\Gamma}^{-1}$
		\STATE \hspace{0.5cm}4. Use $U_{\bm{\Upsilon}}$ to get fine CSI: $U_{\bm{\Upsilon}}\left(\hat{\bm{a}} \right)\longrightarrow \bm{h}_\mathrm{U}$
		\STATE \hspace{0.5cm}5. Compute the loss by $\textbf{(\ref{30})}$
		\STATE \hspace{0.5cm}6. Gradient descent update weight parameters

		\STATE \hspace{0.5cm} $ \textbf{} $

		\STATE\hspace{0.5cm} return $U_{\bm{\Upsilon}}\left(\cdot \right)$		
	\end{algorithmic}
	\label{alg1}
\end{algorithm}

\subsection{Application Level}
At the application level, we mainly focus on tasks of image classification and reconstruction. We will introduce the design of the task-oriented translator, the result aggregation method, the semantic loss function, and the training algorithm for each task, respectively.

\subsubsection{Image Classification Task}

\begin{figure}[htbp]
	\centering
	\includegraphics[width=3.5in]{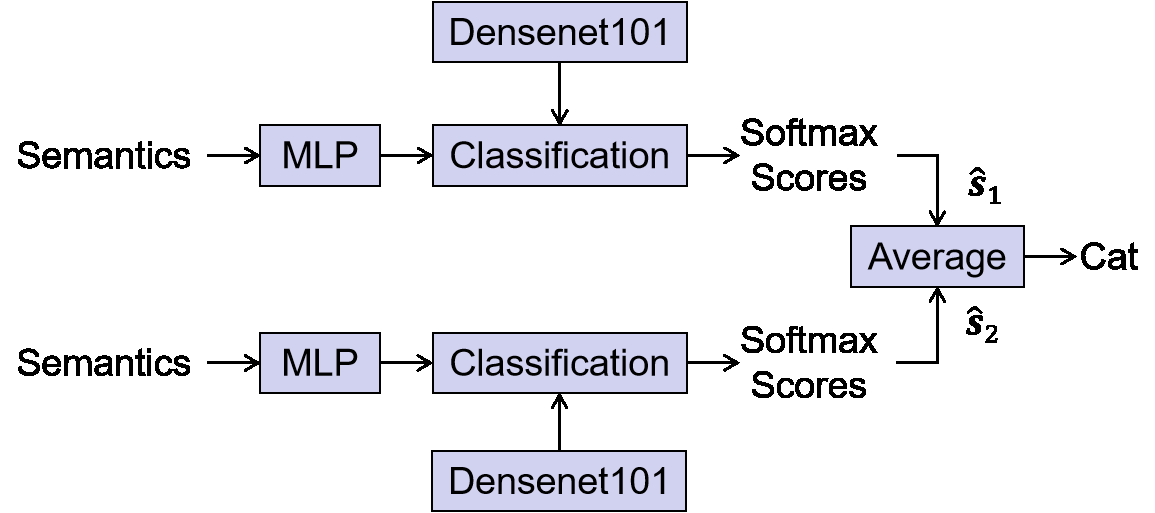}
	\caption{Task-oriented semantic translator for image classification.}
	\label{fig_4}
\end{figure}

\begin{algorithm}[htbp]
	\caption{FLSC Framework for Distributed Image Transmission (Image Classification Task)}\label{alg:alg2}
	\begin{algorithmic}
		\STATE 
		$\textbf{Input:}$ Channel SNR, Device number N, Overlap ratio $\delta$	
		\STATE
		$\textbf{Output:}$ Classification results $\bm{\hat{s}}$, Global model weights $\bm{w}$
		
		
		\STATE
		Load the pretrained U-channel estimator $U_{\bm{\Upsilon}}$
		
		\STATE \hspace{0.5cm}$ \textbf{} $
		
		\STATE1. 
		$\textbf{Server Executes:}$
		\STATE2. \hspace{0.5cm} Initialize $\bm{w}_0$
		\STATE3. \hspace{0.5cm}$ D\gets \vartheta\left(N,\delta\right) $ // get datasets $D$ 
		\STATE4. \hspace{0.5cm}for each device $i = 1,...,N$ do
		\STATE5. \hspace{1cm}give $D_i$ to device $i$ 		
		\STATE6. \hspace{0.5cm}for each round $t = 1,2...,$ do
		\STATE7. \hspace{1cm}for each device $i$ $\mathbf{\in}$ $S_{\mathrm{t}}$ in\, parallel\,do
		\STATE8. \hspace{1.5cm}$\bm{w}_{t+1}^{k}$ $\gets {\textstyle \sum_{k=1}^{K}} \frac{n_{k}}{n}\bm{w}_{t+1}^{k}$
		\STATE9.  \hspace{0.5cm}Aggregate and average all classification results as $\bm{\hat{s}}$	
		\STATE \hspace{0.5cm}$ \textbf{} $
		
		10. $\textbf{LocalUpdate:}$ // Run on device $i$
		\STATE11. \hspace{0.3cm}$\mathbf{\bm{\varrho}}$ $\gets$ (split $D_i$ into batches of size $B$)
		\STATE12. \hspace{0.3cm}for each local epoch $k = 1,...,E$ do
		\STATE13. \hspace{0.8cm}for batch $b$ $\mathbf{\in}$ $\mathbf{\bm{\varrho}}$ do
		\STATE14. \hspace{1.3cm}$ S_{\bm{\alpha}}\left (\bm{s}_{i} \right )\longrightarrow \bm{x}_{i} $
		\STATE15. \hspace{1.3cm}$F_{\bm{\beta}}\left ( \bm{x}_{i} \right )\longrightarrow \bm{y}_{i} $
		\STATE16. \hspace{1.3cm}SVD through the feedback CSI: 
		\STATE \hspace{1.9cm}$\bm{h}_{\mathrm{U}_i}=\bm{U}_i\bm{\Lambda} \bm{V}_{\mathrm{p}_i}^{H}$
		\STATE17. \hspace{1.3cm}$P_{\mathrm{r}}\left (\bm{y}_{i} \right ) \longrightarrow \bm{z}_{i} $
		\STATE18. \hspace{1.3cm}Transmit $\bm{z}_{i}$ over MIMO fading channel: 
		\STATE \hspace{1.9cm}$\bm{\bm{\hat{z}}}_{i}=\bm{h}_{i}\bm{z}_{i}+\bm{n}_{i}$  
		\STATE19. \hspace{1.3cm}$D_{\mathrm{e}}\left (\bm{\bm{\hat{z}}}_{i} \right ) \longrightarrow \bm{\bm{\hat{y}}}_{i} $
		\STATE20. \hspace{1.3cm}Estimate the coarse CSI through LS method: 
		\STATE \hspace{1.9cm}$\bm{h}_{\mathrm{LS}_i}=\hat{\bm{\Gamma}}\bm{\Gamma}^{-1}$
		\STATE21. \hspace{1.3cm}Estimate the feedback fine CSI through $U_{\bm{\Upsilon}}$: 
		\STATE \hspace{1.9cm}$\bm{h}_{\mathrm{U}_i}=U_{\bm{\Upsilon}}\left (\bm{h}_{\mathrm{LS}_i}\right)$
		\STATE22. \hspace{1.3cm}$D_{\bm{\theta}}\left (\bm{\bm{\hat{y}}}_{i}\right )\longrightarrow\bm{\bm{\hat{x}}}_{i}$
		\STATE23. \hspace{1.3cm}$G_{\bm{\eta}}\left (\bm{\bm{\hat{x}}}_{i}\right )\longrightarrow\bm{\hat{s}}_{i}$ // Classfication Results
		\STATE24. \hspace{1.3cm}Compute the loss $L_\mathrm{S}$ by (\ref{10})
		\STATE25. \hspace{1.3cm}Update all the weights $\bm{w}$ for the final $\Psi_{\bm{w}}$                 		
		\STATE \hspace{0.8cm}return $\bm{w}$ to server	
	\end{algorithmic}
	\label{alg2}
\end{algorithm}

The task-specific design is shown in Fig. \ref{fig_4}, in which we use a multi-layer perceptron (MLP) head to solve the image classification task. The MLP head maps the extracted semantic information to the scale of categories. Before generating the classification results, it adapts the knowledge distillation (KD) \cite{Hinton} to obtain the inductive bias as the background knowledge. After the task-oriented translator, each device gets the probabilities for different categories, i.e., the softmax scores. Then, the receiving end collects and averages all the softmax scores, outputting the maximum probable category as the final result.

The KD method \cite{Hinton} is introduced to utilize the prior semantic knowledge in the translator. The final softmax scores are assumed as the soft target to distill its prior semantic knowledge as background information to the student model. In this paper, the teacher model is the Densenet101 \cite{2016Densely}. An extra embedding, called distill embedding \cite{Touvron}, is used to compare the semantic gap between the teacher model and the student model. By integrating the strategy of the KD method, the semantic-aware loss function for image classification becomes

\begin{align}
	\begin{split}
		\label{10}
		L_{\mathrm{S}}=&\left (1-\lambda \right)L_{\mathrm{CE}}\left (\varPhi\left(\log\left (\hat{\bm{s}}\right)\right),\varphi\left(\bm{s}\right)\right)\\+
		&\lambda \tau_{\mathrm{K}}^{2}D_{\mathrm{KL}}\left (\varPhi\left(\log\left (\hat{\bm{s}}\right)/\tau_{\mathrm{K}} \right),\varPhi \left(\log\left (S_{\mathrm{T}}\left(\bm{s},\bm{w}_{\mathrm{T}}\right)\right)/\tau_{\mathrm{K}}\right)\right),  
	\end{split}
\end{align}
where $L_{\mathrm{CE}}(\cdot,\cdot)$ is the common cross-entropy loss that measures the difference between student target, $\varPhi\left(\log\left(\hat{\bm{s}}\right)\right)$, and true target, $\varphi\left(\bm{s}\right)$. The second part, $D_{\mathrm{KL}}(\cdot,\cdot)$, is the Kullback-Leibler (KL)-divergence \cite{KL} that compares the distribution of the student target with the teacher target. Parameter $\lambda$ is the trade-off for $L_{\mathrm{CE}}(\cdot,\cdot)$ and $D_{\mathrm{KL}}(\cdot,\cdot)$. $\tau_{\mathrm{K}}$ is the distillation temperature that controls the speed of distillation procedure. $S_{\mathrm{T}}(\cdot,\cdot)$ and $\bm{w}_{\mathrm{T}}$ are the networks and parameters of teacher model Densenet101, respectively. In this way, we can transfer the semantic information from the teacher model to the semantic communication network.

We train the FLSC in an end-to-end manner with the semantic-aware FL algorithm. According to different SNRs, corresponding pretrained U-channel estimated is first loaded into the FLSC. After that, the U-channel estimator trains with the whole network together. Training algorithm for the image classification task is shown in Algorithm \ref{alg2}, where we assume the overlap ratio among different images as $\delta$ and function $\vartheta(\cdot,\cdot)$ for data allocation.

\subsubsection{Image Reconstruction Task}

\begin{figure}[htbp]
	\centering
	\includegraphics[width=3.5in]{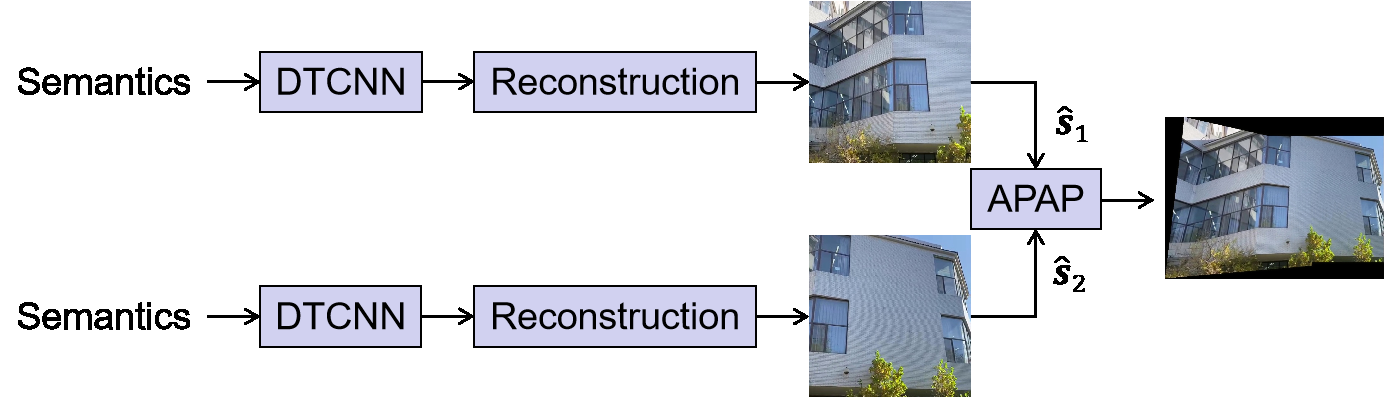}
	\caption{Task-oriented semantic translator for image reconstruction.}
	\label{fig_5}
\end{figure}

We assume $G_{\bm{\eta}}(\cdot)$ as the DTCNN for image reconstruction. As in Fig. \ref{fig_5}, the DTCNN reconstructs the semantic information into the original images. After getting the reconstructed images from each device, we collect all the images and stitch them to a panorama that contains all the features from each image. As-Projective-As-Possible Image Stitching (APAP) \cite{APAP} is adopted for the image stitching in our result aggregation. It estimates the transformation function and matching points which bring different images into alignment, then mapping all the aligned images onto a common canvas as a panorama.

Moreover, for semantic-aware image reconstruction, we utilize the mean square error (MSE) as the loss function, which is defined as
\begin{align}
	\label{13}
	L_{\mathrm{S}}=\sum_{i=0}^{H}\sum_{j=0}^{W}\sum_{k=0}^{C}||(\varphi\left(\bm{s}\right))_{i,j,k}-(\hat{\bm{s}})_{i,j,k}||^2,     
\end{align}
where the image size is \textit{H$\times$W$\times$C}.

\begin{algorithm}[htbp]
	\caption{FLSC Framework for Distributed Image Transmission (Image Reconstruction Task)}\label{alg:alg3}
	\begin{algorithmic}
		\STATE 
		$\textbf{Input:}$ Channel SNR, Device number N, Overlap ratio $\delta$		
		
		$\textbf{Output:}$ Stitched images $\bm{\hat{s}}$, Global model weights $\bm{w}$
		
		
		\STATE
		Load the pretrained U-channel estimator $U_{\bm{\Upsilon}}$
		
		\STATE \hspace{0.5cm}$ \textbf{} $
		
		\STATE1. 
		$\textbf{Server Executes:}$
		\STATE2. \hspace{0.5cm}Initialize $\bm{w}_0$
		\STATE3. \hspace{0.5cm}$ D\gets \vartheta\left(N,\delta\right) $ // get datasets $D$ 
		\STATE4. \hspace{0.5cm}for each device $i = 1,...,N$ do
		\STATE5. \hspace{1cm}give $D_i$ to device $i$ 		
		\STATE6. \hspace{0.5cm}for each round $t = 1,2...,$ do
		\STATE7. \hspace{1cm}for each device $i$ $\mathbf{\in}$ $S_{\mathrm{t}}$ in\, parallel\,do
		\STATE8. \hspace{1.5cm}$\bm{w}_{t+1}^{k}$ $\gets {\textstyle \sum_{k=1}^{K}} \frac{n_{k}}{n}\bm{w}_{t+1}^{k}$
		\STATE9. \hspace{0.5cm}Aggregate all the reconstructed images and 
		\STATE \hspace{0.9cm}concatenate them with APAP as $\bm{\hat{s}}$
		\STATE10. \hspace{0.3cm}Evaluate the reconstructed and stitched images by 
		\STATE \hspace{0.9cm}$\bm{s}_{\mathrm{sum}}$ and $\bm{\hat{s}}$		
		
		\STATE \hspace{0.5cm}$ \textbf{} $
		
		11. $\textbf{LocalUpdate:}$ // Run on device $i$ 
		\STATE12. \hspace{0.3cm}$\mathbf{\bm{\varrho}}$ $\gets$ (split $D_i$ into batches of size $B$)
		\STATE13. \hspace{0.3cm}for each local epoch $k = 1,...,E$ do
		\STATE14. \hspace{0.8cm}for batch $b$ $\mathbf{\in}$ $\mathbf{\bm{\varrho}}$ do
		\STATE15. \hspace{1.3cm}$ S_{\bm{\alpha}}\left (\bm{s}_{i} \right )\longrightarrow \bm{x}_{i} $
		\STATE16. \hspace{1.3cm}$F_{\bm{\beta}}\left ( \bm{x}_{i} \right )\longrightarrow \bm{y}_{i} $
		\STATE17. \hspace{1.3cm}SVD through the feedback CSI: 
		\STATE \hspace{1.9cm}$\bm{h}_{\mathrm{U}_i}=\bm{U}_i\bm{\Lambda} \bm{V}_{\mathrm{p}_i}^{H}$
		\STATE18. \hspace{1.3cm}$P_{\mathrm{r}}\left (\bm{y}_{i} \right ) \longrightarrow \bm{z}_{i} $
		\STATE19. \hspace{1.3cm}Transmit $\bm{z}_{i}$ over MIMO fading channel: 
		\STATE \hspace{1.9cm}$\bm{\bm{\hat{z}}}_{i}=\bm{h}_{i}\bm{z}_{i}+\bm{n}_{i}$  
		\STATE20. \hspace{1.3cm}$D_{\mathrm{e}}\left (\bm{\bm{\hat{z}}}_{i} \right ) \longrightarrow \bm{\bm{\hat{y}}}_{i} $
		\STATE21. \hspace{1.3cm}Estimate the coarse CSI through LS method: 
		\STATE \hspace{1.9cm}$\bm{h}_{\mathrm{LS}_i}=\hat{\bm{\Gamma}}\bm{\Gamma}^{-1}$
		\STATE22. \hspace{1.3cm}Estimate the feedback fine CSI through $U_{\bm{\Upsilon}}$: 
		\STATE \hspace{1.9cm}$\bm{h}_{\mathrm{U}_i}=U_{\bm{\Upsilon}}\left (\bm{h}_{\mathrm{LS}_i}\right)$
		\STATE23. \hspace{1.3cm}$D_{\bm{\theta}}\left (\bm{\bm{\hat{y}}}_{i}\right )\longrightarrow\bm{\bm{\hat{x}}}_{i}$
		\STATE24. \hspace{1.3cm}$G_{\bm{\eta}}\left (\bm{\bm{\hat{x}}}_{i}\right )\longrightarrow\bm{\hat{s}}_{i}$    
		// Reconstruction Images
		\STATE25. \hspace{1.3cm}Compute the loss $L_\mathrm{S}$ by (\ref{13})
		\STATE26. \hspace{1.3cm}Update all the weights $\bm{w}$ for the final $\Psi_{\bm{w}}$                		
		\STATE \hspace{0.8cm}return $\bm{w}$ to server		
	\end{algorithmic}
	\label{alg3}
\end{algorithm}

Training algorithm for image reconstruction task is shown in Algorithm \ref{alg3}.

\section{Numerical Results}
In this section, numerical results of the proposed FLSC framework are provided. We introduce the simulation setups first. After that, the experiments of the FLSC in terms of image classification and reconstruction in comparison with different competitors are provided. Some visualized results and real-scene examples are also given in order to give more intuitive comparison for the FLSC. Then, the superiority of our proposed FLSC versus the central learning schemes is demonstrated. Finally, we show the effectiveness of the hierarchical structures in terms of different tasks.

\subsection{Experimental Setups}
$\textbf{Datasets:}$
We consider the distributed iamge transmission with IoT devices for $N=2$, and assume that each local device with different MIMO fading channels experiences the same SNR. The MIMO channel matrix varies batches by batches following block fading. The number of antennas in the MIMO system is $N_{\mathrm{T}}$ = 2 and $N_{\mathrm{R}}$ = 2, respectively. MNIST and CIFAR10 datasets are used for image classification while an extra dataset, namely UDIS-D \cite{UDIS-D}, is implemented for image reconstruction task. The UDIS-D is a comprehensive real-world dataset, which contains common scenes, such as the car park, grass field, building, etc. It has 10,440 pairs of images for training and 1,106 pairs for testing, where images are shot in different angles, illumination, and overlap ratios. Three-level overlap ratio catagories are defined, where high overlap ratio $\delta$ is greater than 90$\%$, middle overlap ratio ranges from 60$\%$ to 90$\%$, and low overlap rate is lower than 60$\%$.

$\textbf{Metrics:}$
We evaluate the classification accuracy for different datasets and schemes in image classification. For image reconstruction and stitching, we use the peak signal-to-noise ratio (PSNR) and structural similarity (SSIM) as measurements for the quality of reconstructed and stitched images.

$\textbf{Control parameters:}$
In our experiments, we mainly focus on two parameters, namely the image overlap ratio $\delta$ and the channel bandwidth ratio $R$. The image overlap ratio, $\delta$, is determined by the proportion of overlapping areas between two images and reflects the difference of angles and distances among several cameras in the distributed scenario. The channel bandwidth ratio, $R$, measures the compressed extent from the raw image to the transmitted semantic information sequence, namely $R=\frac{C_{\mathrm{L}}}{H\times W\times C}$.

$\textbf{Training details:}$
We use the variable learning rate, which decreases step-by-step from 0.0005 to 0.0001. The optimization algorithm is AdamW and the weight decay is 0.05. We use the labelsmoothing \cite{szegedy2016rethinking} strategy to alleviate the overfitting and the labelsmoothing weight $\mu$ is 0.2. The distill temperature, $\tau_{\mathrm{K}}$, in KD is 0.5. The FL framework trains 40 global rounds for image classification and 30 global rounds for image reconstruction, while each local device trains 10 local rounds for both tasks. Our HVT basically has 4 layers with depths 2, 2, 2, 2 and output channel numbers 64, 128, 320, 512, respectively. Two layers are used for image classification while four layers are used for image reconstruction in the HVT. For simplicity, we resize the images of MNIST into 32$\times$32$\times$1 and the UDIS-D into 128$\times$128$\times$3.

$\textbf{Baseline:}$
We use the vision transformer (ViT), deep joint source and channel coding (JSCC), and joint photographic experts group (JPEG) combined with low-density parity-check (LDPC) as the benchmarks. The ViT is set according to the ViT-small model in \cite{Dosovitskiy} for image classification. We adopt the JSCC scheme in \cite{JSCC} for image reconstruction, while adding a MLP head for image classification. JPEG+LDPC1/2 functions as the traditional separated source and channel coding method, where the channel coding rate is 1/2. For image reconstruction, the FLSC and JSCC schemes obey the three-level overlap ratio catagories of UDIS-D, while JPEG+LDPC1/2 method considers all the images.

\subsection{Image Classification}
For the image classification task, we utilize the MNIST and CIFAR10 as the datasets, and
explore the performance on classification accuracy over different key quantities, e.g. SNRs and image
overlap ratios $\delta$.

\begin{figure}[htbp]
	\centering
	\includegraphics[width=3.6in]{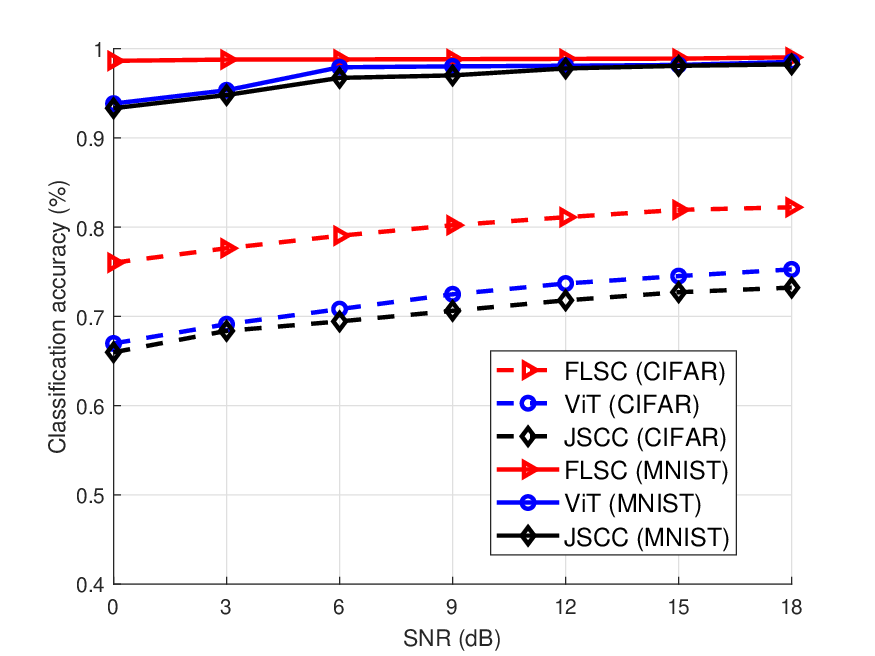}
	\caption{The classification accuracy of different schemes versus the SNRs in MIMO fading channels ($\delta$ = 0.6, $R$ = 0.04).}
	\label{fig_7}
\end{figure}

Fig. \ref{fig_7} shows the classification accuracy versus the SNRs for the FLSC, ViT, and JSCC, with a moderated overlapping ratio $\delta$ = 0.6 and channel bandwidth ratio $R$ = 0.04. For both datasets, our FLSC surpasses other benchmarks in terms of SNRs ranging from 18 dB to 0 dB, especially when SNRs are low. The accuracy also drops slowly as SNR decreases compared to JSCC and ViT, which certificates the noise robustness of the FLSC. It is worth mentioning that ViT has a better anti-noise performance than the JSCC method since it introduces the attention mechanism instead of CNNs to extract semantics.

\begin{figure}[htbp]
	\centering
	\includegraphics[width=3.6in]{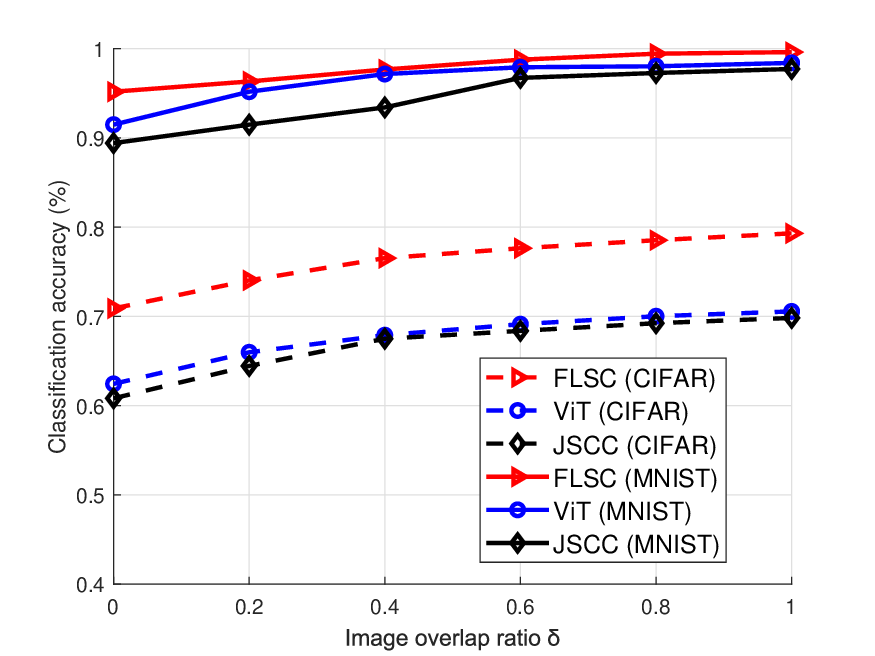}
	\caption{The classification accuracy of different schemes versus the image overlap ratios in MIMO fading channels (SNR = 3 dB, $R$ = 0.04).}
	\label{fig_8}
\end{figure}

Fig. \ref{fig_8} shows the classification accuracy versus the image overlap ratio, $\delta$, where SNR = 3 dB. For all schemes, the accuracy decreases with the drop of $\delta$. This is because that more overlap area of images can help the model better extract needed semantic information. Nevertheless, the FLSC outperforms ViT and JSCC since it not only has good accuracy with high overlap ratio but also decreases relatively slow as the overlap ratio decreases. Even in the completely non-overlapped condition ($\delta$ = 0), the FLSC still owns reasonably good performances. 

\begin{figure*}[htbp]
	\centering  
	\subfigure[PNSR for the reconstructed images.]{
		\includegraphics[width=0.44\linewidth]{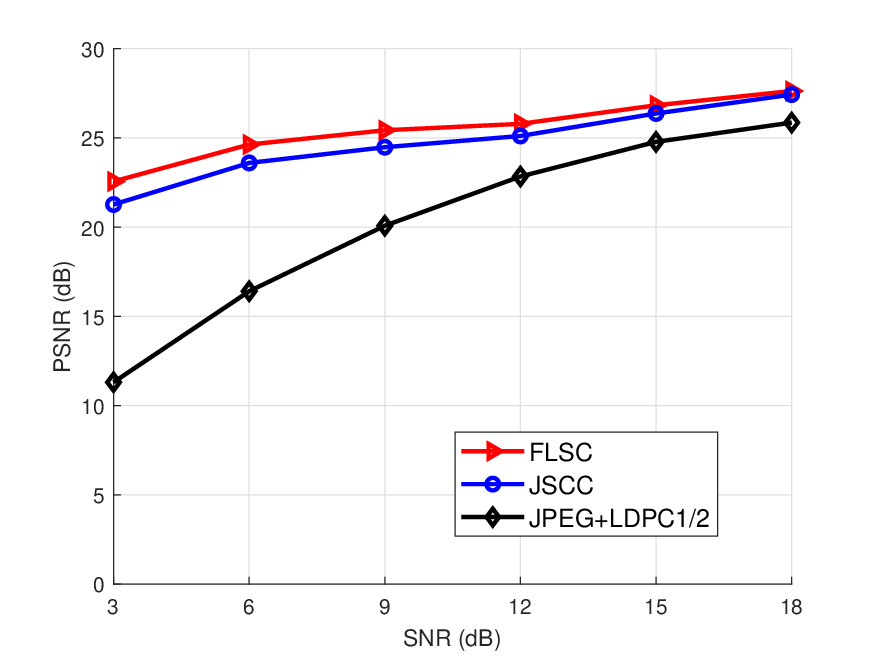}}
	\subfigure[SSIM for the reconstructed images.]{
		\includegraphics[width=0.44\linewidth]{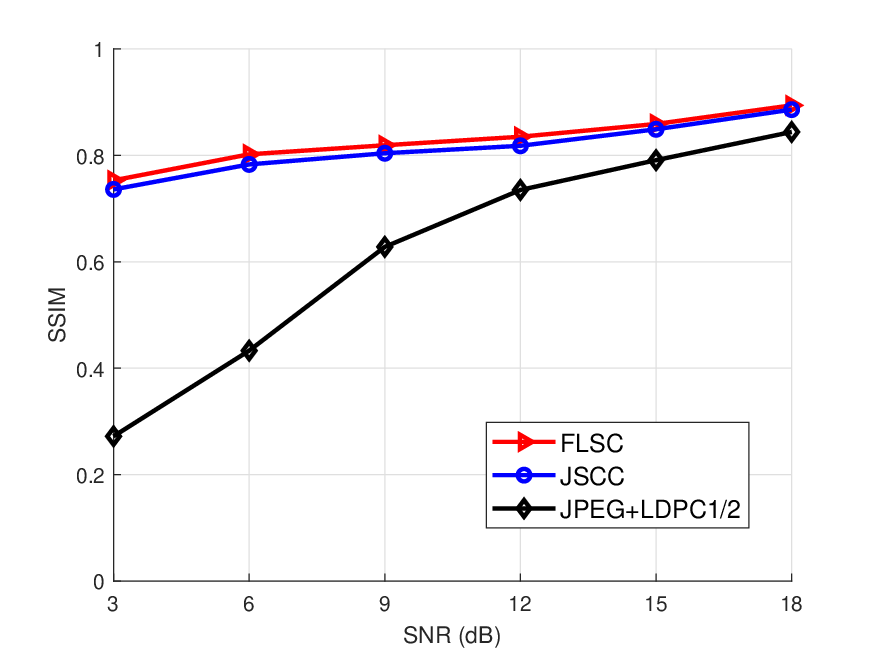}}
	\caption{Quality of the reconstructed images versus the SNRs in MIMO fading channels ($R$ = 0.08).}
	\label{fig_10}
\end{figure*}

\begin{figure*}[htbp]
	\centering  
	\subfigure[PNSR for the reconstructed images.]{
		\includegraphics[width=0.44\linewidth]{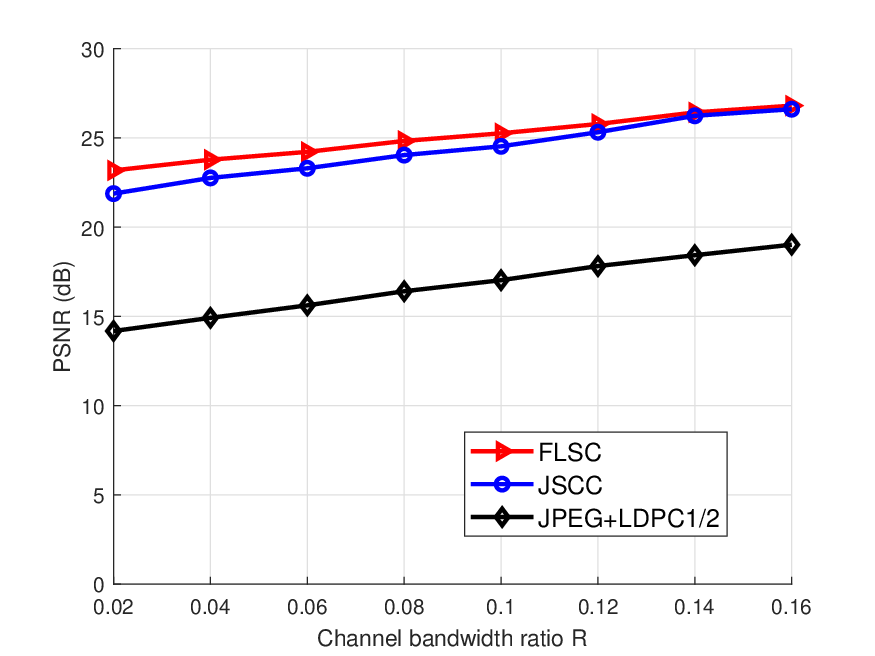}}
	\subfigure[SSIM for the reconstructed images.]{
		\includegraphics[width=0.44\linewidth]{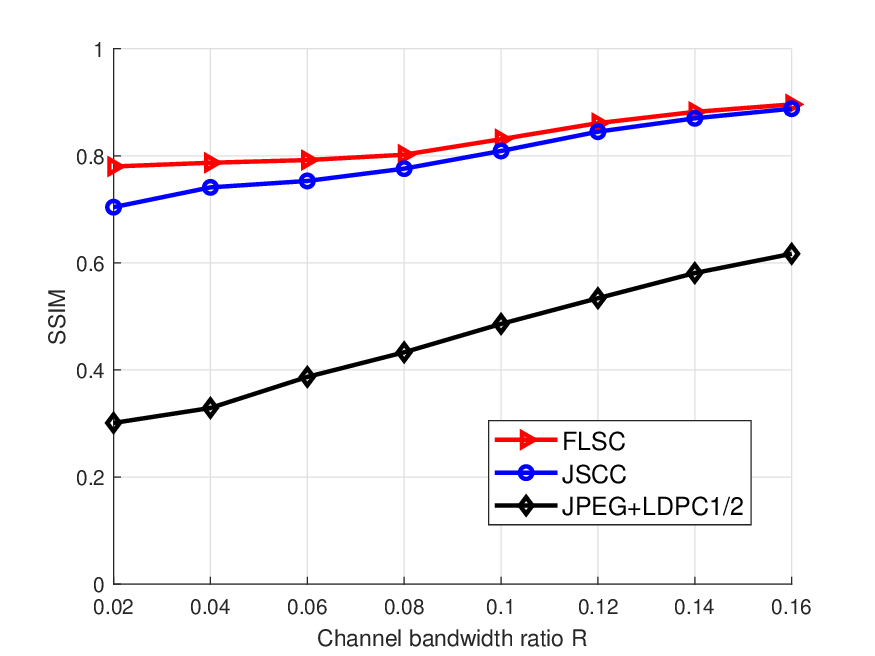}}
	\caption{Quality of the reconstructed images versus the channel bandwidth ratios in MIMO fading channels (SNR = 6 dB).}
	\label{fig_11}
\end{figure*}

\begin{figure*}[htbp]
	\centering  
	\subfigure[PNSR for the stitched images.]{
		\includegraphics[width=0.44\linewidth]{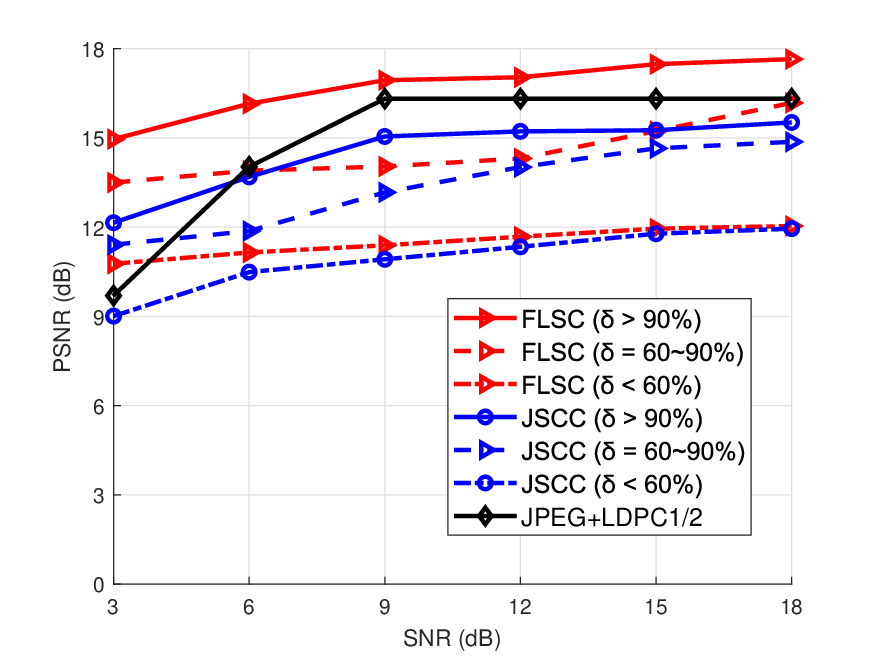}}
	\subfigure[SSIM for the stitched images.]{
		\includegraphics[width=0.44\linewidth]{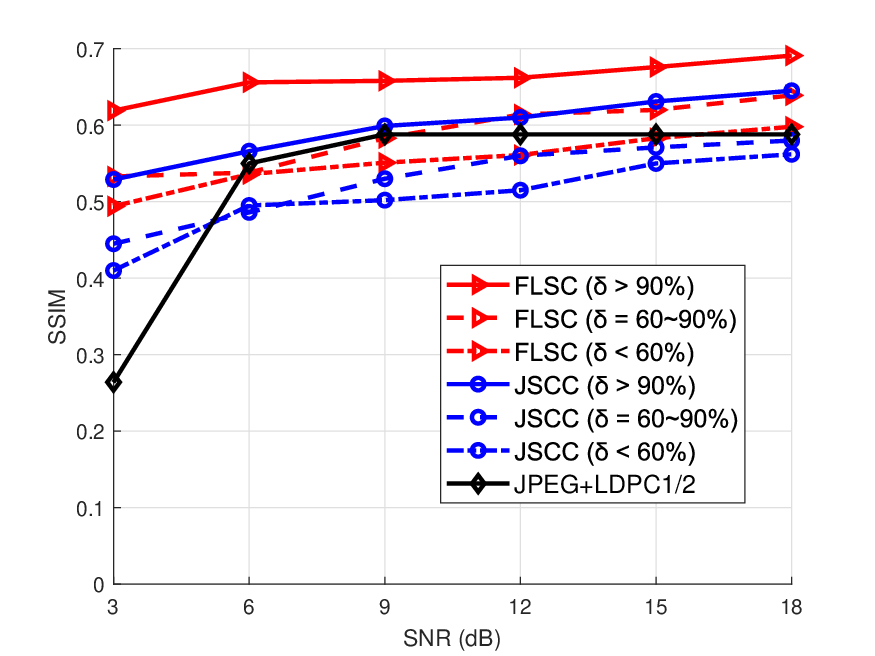}}
	\caption{Quality of the stitched images in MIMO fading channels.}
	\label{fig_12}
\end{figure*}

\begin{figure*}[htbp]
	\centering
	\includegraphics[width=6.4in]{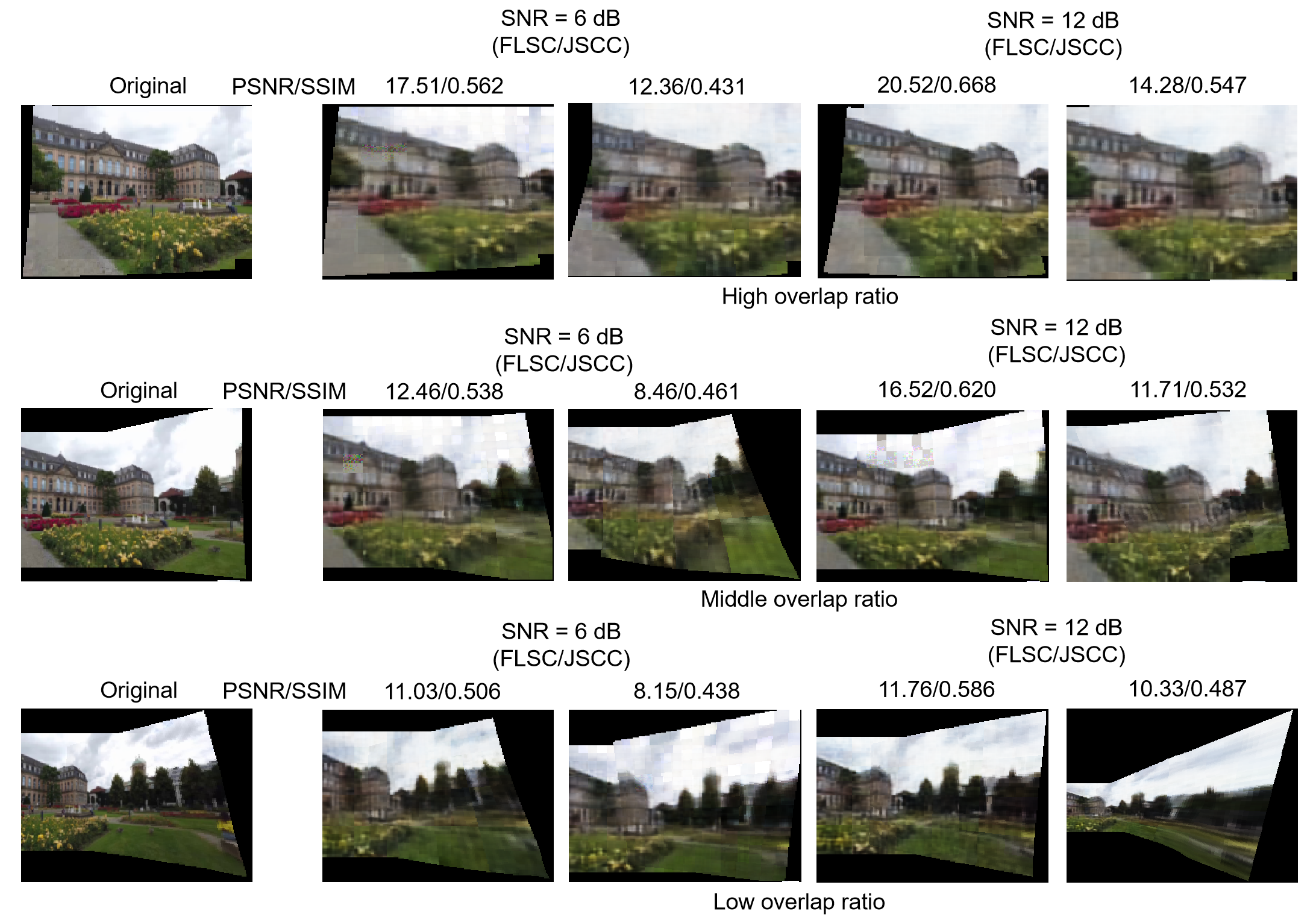}
	\caption{Visualization for stitched images of three-level overlap ratio, where PSNR and SSIM are given. SNRs are 6 and 12 dB, and the methods of FLSC and JSCC are considered.}
	\label{fig_13}
\end{figure*}

\subsection{Image Reconstruction}
For the image reconstruction task, we first consider the quality of reconstructed images through FLSC and other competitors using the UDIS-D dataset. Then we evaluate the PSNR and SSIM of stitched images which are concatenated from reconstructed images of all the local devices. Finally, some visualized results and an extra real-scene example are given.

Fig. \ref{fig_10} illustrates the PSNR and SSIM of the reconstructed images versus different SNRs in MIMO fading channel. Compared to the other two candidates, the FLSC produces the best image quality in terms of PSNR and SSIM with different SNR regimes. Confronted with the MIMO fading, the AI-based methods outperform the traditional source and channel coding method such as JPEG+LDPC1/2, especially in low SNR conditions. Even faced with nonideal or extremely worse channel conditions, eg. SNR = 3 dB , FLSC is equipped to acquire reasonably satisfying transmission results.  

Fig. \ref{fig_11} presents the PSNR and SSIM performances versus the channel bandwidth ratios, $R$, where SNR = 6 dB. It illustrates the bandwidth usage of each scheme since $R$ reflects the average sum of bits to be transmitted per image. For a large range of channel bandwidth ratios, eg. $R$ ranges from 0.02 to 0.16, FLSC consistently outperforms the AI-based JSCC method. For the separation-based methods such as JPEG+LDPC, the performance gap is even enlarged compared to the JSCC method. Indeed, when it comes to different evaluation metrics, the performances of FLSC surpass other schemes both in PSNR and SSIM, which means that human perception as well as accurate reconstruction can be well conducted under different bandwidth usage conditions. Overall, FLSC shows competitive performances with different bandwidth usages. Considering the extreme channel bandwidth ratio condition of $R = 0.02$, FLSC is still able to reconstruct the images with good quality relatively.

Fig. \ref{fig_12}(a) and Fig. \ref{fig_12}(b) show the PSNR and SSIM performances of stitched images from the reconstructed images versus SNRs. The stitched images seem to have the similar trend like the reconstructed images, in which the FLSC not only surpasses JSCC but also outperforms JPEG+LDPC1/2 from SNR 18 dB to 3 dB. As other schemes ignore the semantic information from the raw images, the FLSC extracted needed semantics which help the APAP method better find the matching points to produce the full panorama. In this way, in different levels of the image overlap ratio, the FLSC is able to present reasonably good-quality stitched images.

Then, we give some visualized results for the stitched images from the UDIS-D dataset with different $\delta$ and SNR cases, as shown in Fig. \ref{fig_13}. The FLSC can aggregate each reconstructed image to a full panorama with different overlap ratio and SNR regimes. The semantic information retained in the reconstructed images ensures the APAP method to better stitch the final panorama. It also provides more anti-noise effect since the semantics in images are harder to be confused compared to the usual bit recovery schemes. Unlike the JSCC which has blurry and incorrect stitched results, our FLSC has more stability in terms of distributed image transmission tasks.

\begin{figure}[htbp]
	\centering
	\includegraphics[width=3.3in]{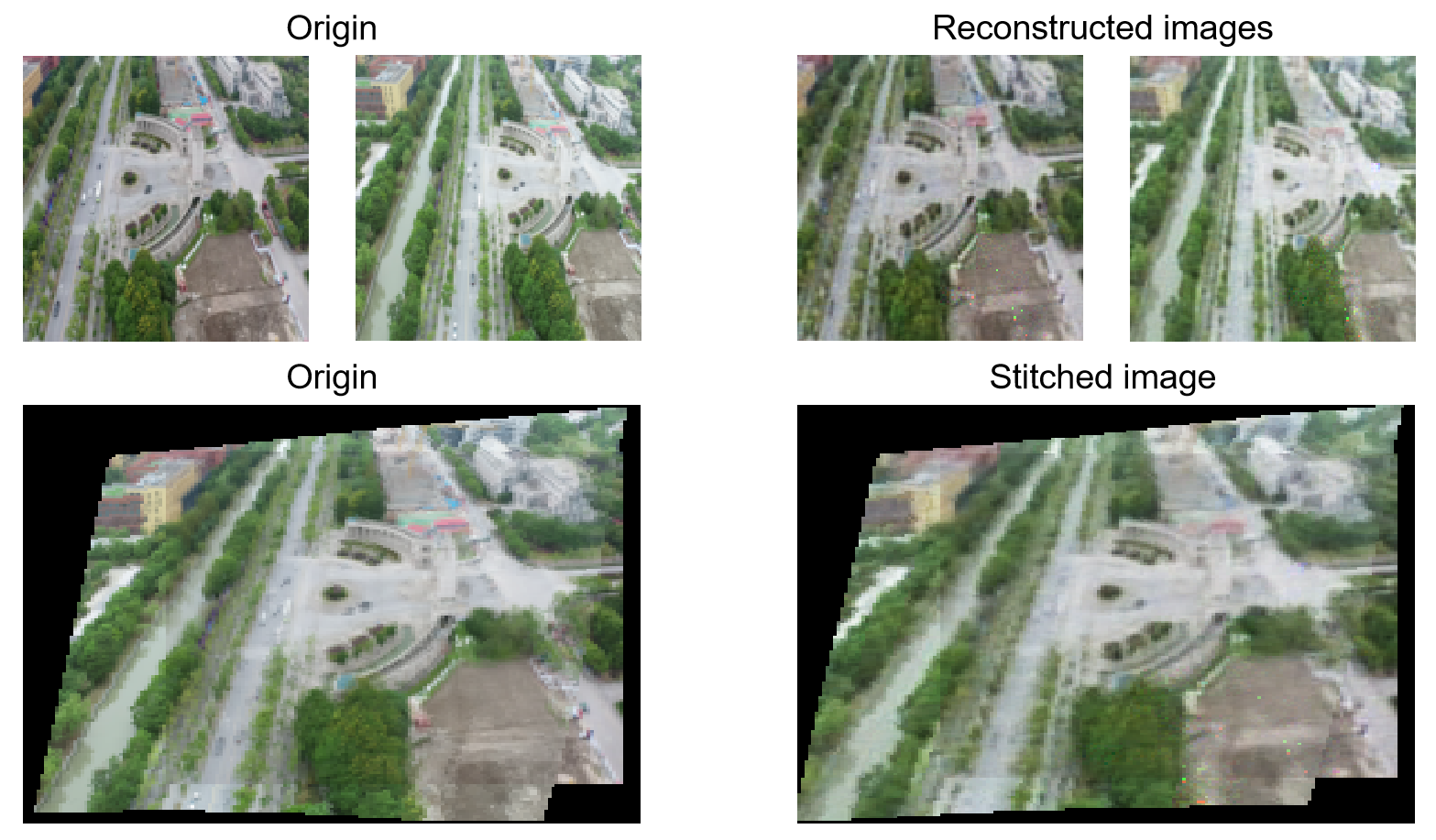}
	\caption{Real scene example for the reconstructed images and stitched images.}
	\label{fig_14}
\end{figure}

Finally, we give some real examples to test the practicability and generalization ability of the FLSC. We use two UAVs and shoot two surveillance videos of the same region from different angles \cite{DVDET}. As shown in Fig. \ref{fig_14}, the original images, reconstructed images, and stitched images are given, respectively. Obviously, the original image pair not only has different angles but also different illumination as well, which may cause extra difficulties for the distributed image transmission. However, for our stitched images, the illumination is kept as a unified state for the full panorama and the scene is nearly the same as the raw stitched image as well.

\subsection{The Performances of the FL Schemes}

\begin{figure}[htbp]
	\centering
	\includegraphics[width=3.6in]{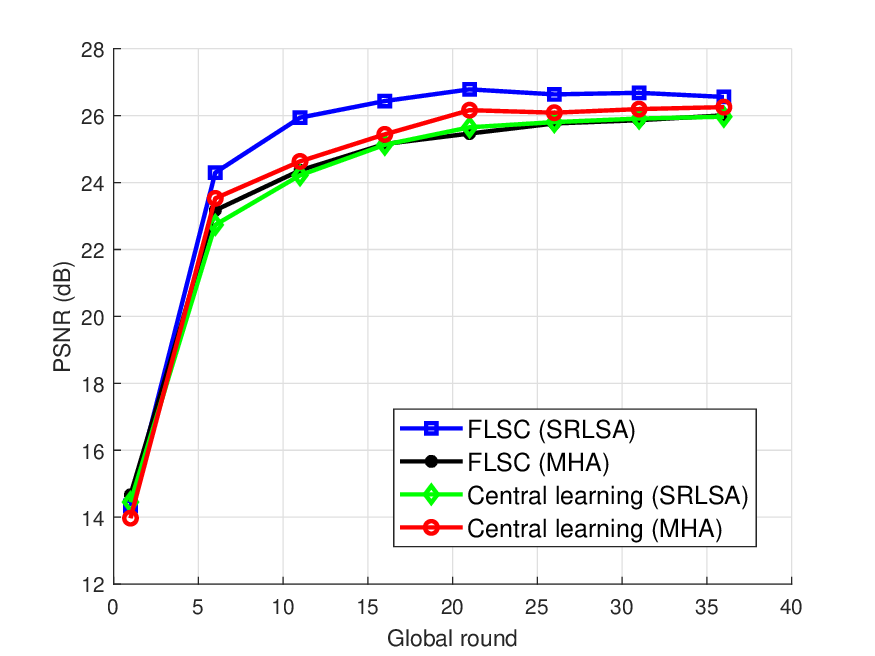}
	\caption{Performances of the FLSC compared to the central learning.}
	\label{fig_15}
\end{figure}

Considering the distributed scenario, we compare our FLSC to the conventional central learning schemes. Here, central learning means that all source data are trained and tested through a point-to-point communication link just as the single-user one in FLSC.

Since FL is an iterative process, similar to [20], the PSNR results versus the required global rounds for the FLSC and central learning is shown in Fig. \ref{fig_15} to evaluate the computation cost and the communication cost. Among them, SRLSA is our proposed method in HVT, while MHA is the intrinsic module embedded in the transformer. We can observe that the FLSC achieves better results than the central learning ones, while requiring a lesser number of global rounds to train the framework acquiring equivalent performance compared to central learning approaches. Then, with the proposed SRLSA, the final results are further improved compared with the ones with original MHA modules. These results serve as evidences of the reduced communication and computation cost associated with the FLSC compared to other schemes. Furthermore, since FL is a distributed method, the total computation cost can be splitted among different users, thus alleviating the overhead for each individual user. The better final results further demonstrate the less communication and computation cost for the FLSC.   

\subsection{The Effectiveness of the Hierarchical Structure}

\begin{figure}[h]
	\centering  
	\subfigure[For image classification (CIFAR).]{
		\includegraphics[width=0.85\linewidth]{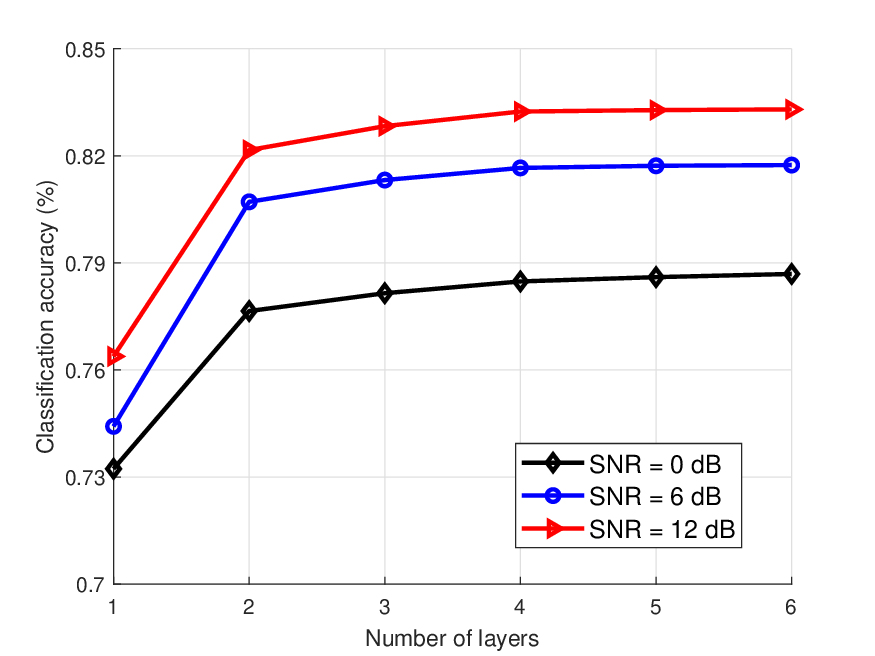}}
	\\
	\subfigure[For image reconstruction (SNR = 10 dB).]{
		\includegraphics[width=0.54\linewidth]{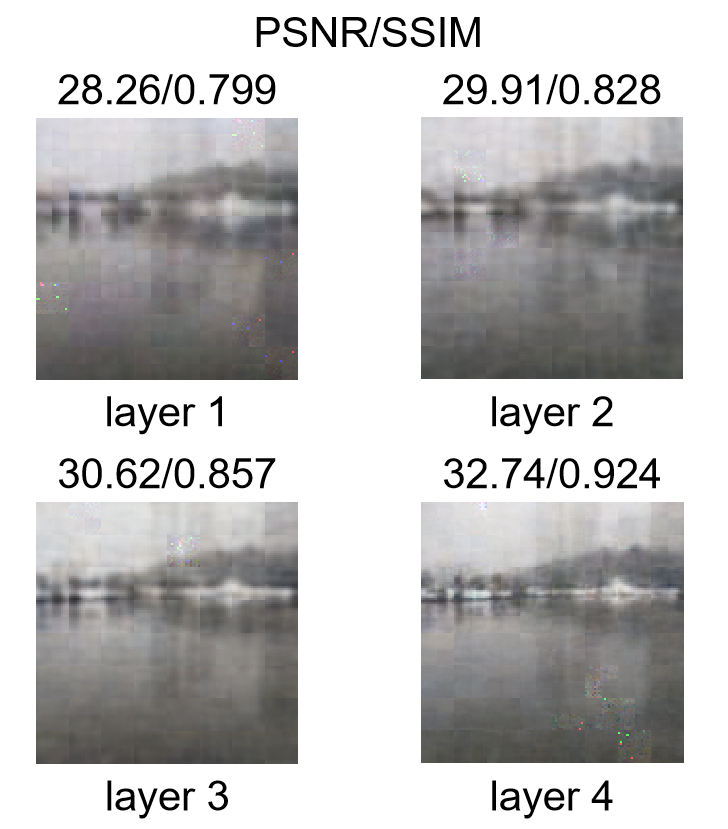}}
	\caption{Performances of hierarchical structures in terms of different tasks.}
	\label{fig_16}
\end{figure}

In Fig. \ref{fig_16}, we present the performance of different layer numbers of HVT for each task. For Fig. \ref{fig_16}(a), we can find that our HVT with only one layer has the worst performance compared to others. When with two layers, the classification accuracy becomes much better. However, if three layers are given, the performance only increases a little while with four or more layers the performance stays relatively the same. Consider the sum of transmitted semantics and final classification accuracy as a trade-off, the relatively lightweight two-layers structure is suitable for image classification task. In Fig. \ref{fig_16}(b), the PSNR and SSIM increase continuously with the increase of layer numbers. We draw the conclusion that deeper network structure of HVT enhances the performance of image reconstruction, but merely the first two layers are required for image classification task. Therefore, a flexible HVT with variable layers can be designed for different tasks in the proposed FLSC framework.

\section{Conclusion}
This paper proposed a communication-efficient semantic communication framework FLSC for distributed image transmission with IoT devices. It can be interpreted into three levels: semantic level adapts HVT and DTCNN to extract and recover the semantic information for communication overhead reduction. Transmission level conducts SVD for MIMO precoding and detection based on CSI while introducing a two-step channel estimation method to acquire fine CSI. Application level takes the task-related design into consideration and introduces the task-oriented translator, result aggregation method, semantic loss function, and training method for the image classification and image reconstruction tasks, respectively. Besides, we have performed a series of experiments for the FLSC and other benchmarks. Results show the superiority of our FLSC over other deep-learning based method such as JSCC and traditional source and channel coding method such as JPEG+LDPC especially in conditions of limited bandwidth resources, low overlap ratios, and low SNRs. Even in extreme compress ratio such as 0.02 or low SNR such as 3 dB, FLSC still performs different tasks well compared to other benchmarks. Real-scene examples and visualized results further demonstrate the generalization and practicality of the FLSC. The effectiveness of the hierarchical structure has also been verified through the trade-off of task performances and computation complexity. In summary, this paper proposed a promising framework to enable the task adaptation and efficient image transmission under the distributed scenarios. In the future, emerging heterogeneous multi-media sources among different users will be taken into consideration for FLSC. Also, new criteria for evaluating the performances of semantic tasks in distributed transmission will also be studied.

\fontsize{8pt}{10pt}\selectfont

\bibliography{ref}

\vfill

\end{document}